\documentclass[11pt]{article}

\usepackage[preprint]{acl}
\usepackage{iftex}
\ifPDFTeX
  \usepackage[T1]{fontenc}
  \usepackage[utf8]{inputenc}
  \usepackage{tgtermes}
  \usepackage{tgcursor}
  \newcommand{\datasetfont}{\fontfamily{lmtt}\selectfont}
\else
  \usepackage{fontspec}
  \newfontfamily\datasetfont{Latin Modern Mono}
\fi
\usepackage{soul}
\usepackage{tabularx}

 \usepackage{microtype}
 \usepackage{booktabs} 
\usepackage{amsmath}
\usepackage{amssymb}
\usepackage{dsfont}
\usepackage{nccmath}
\usepackage{graphicx}
 \usepackage{multirow}
 \usepackage{subcaption}
 \usepackage[most]{tcolorbox}
\usepackage{tikz}
\definecolor{lightgreen}{RGB}{220,245,220}
\definecolor{darkgreen}{RGB}{0,100,0}

\usepackage{pifont}
\usepackage{xspace}
\usepackage{xcolor}
\usepackage{enumitem}

\usepackage{booktabs, multirow, array, makecell}
\usepackage{graphicx}

\newcolumntype{P}[1]{>{\centering\arraybackslash}p{#1}}
\newcolumntype{L}[1]{>{\raggedright\arraybackslash}p{#1}}

\newcommand{\benchmark}{\textsc{IIF-Bench}\xspace}
\newcommand{\method}{\textsc{Graft}\xspace}
\newcommand{\sft}{\textsc{Graft-SFT}\xspace}
\newcommand{\vrpo}{\textsc{Graft-DPO}\xspace}
\newcommand{\llmname}[1]{\texttt{\hyphenchar\font=\defaulthyphenchar #1}\xspace}
\newcommand{\datasetname}[1]{{\datasetfont #1}\xspace}
\usepackage{colortbl} 
\definecolor{modelshade}{RGB}{242,246,250}
\definecolor{gaincolor}{RGB}{24,128,88}
\definecolor{vanillagray}{RGB}{90,90,90}

\newcommand{\gain}[1]{\textcolor{gaincolor}{\raisebox{-0.35ex}{\tiny$\mathrm{\uparrow\!\!#1}$}}}
\newcommand{\vanilla}[1]{\textcolor{vanillagray}{#1}}

\definecolor{modelshade}{RGB}{238,244,250}

\newlength{\modelrowwidth}
\newcommand{\modelrow}[1]{\multicolumn{9}{c}{\cellcolor{modelshade}\textbf{#1}}}

\title{In-Place Instruction Following in Diffusion Language Models}

\author{
 \textbf{Zheng Nie\textsuperscript{1,*}},
 \textbf{Zherui Li\textsuperscript{2,*}},
 \textbf{Jiaming Zhang\textsuperscript{2}},
 \textbf{Kun Wang\textsuperscript{2}},
 \\
 \textbf{Zhenhong Zhou\textsuperscript{2,$\dagger$}},
 \textbf{Yufei Guo\textsuperscript{3,$\dagger$}}
\\
  \textsuperscript{1}National University of Singapore
  \\
 \textsuperscript{2}Nanyang Technological University
 \\
 \textsuperscript{3}Peking University\\
}

\AddToHook{cmd/@trivlist/after}{\setlength{\rightskip}{0pt plus 1em}}

\begin{document}
\setlength{\rightskip}{0pt plus 1em}
\raggedbottom

\maketitle

\begin{abstract}
Diffusion Large Language Models (dLLMs) generate text via bidirectional iterative denoising, naturally supporting user-specified constraints anchored at arbitrary output positions, a paradigm known as In-place Prompting (IPP). We formalize this as the In-place Instruction Following (IIF) task and construct \benchmark, a hierarchical benchmark spanning literal, style, and discourse-function constraints, paired with a rubric-based local-global evaluation protocol. An inference-time attention-bias probe suggests that vanilla dLLMs often under-prioritize constraint spans during denoising. We then propose \method, an IPP-oriented post-training framework combining constraint-aware SFT and preference optimization. On four representative dLLMs, \method raises the average IIF score from 57.75 to 73.10 ($\uparrow$15.35 points), with absolute gains of 15.91 and 15.57 points on literal and discourse-function constraints, while preserving general generation ability. Our code and data are available at: \url{https://anonymous.4open.science/r/Graft-36B5}.
\end{abstract}

\section{Introduction}


The rapid development of Large Language Models (LLMs) has significantly enhanced the instruction-following capabilities of natural language generation systems~\citep{gpt2, llm-survey, zhou2023instruction, if-survey}. 
Currently, mainstream LLMs still primarily rely on prefix-based prompting as their mode of interaction: users specify task requirements~\citep{gpt3, multitask}, format constraints~\citep{liu2024we, dllmcd}, or preference descriptions~\citep{li2024dissecting, zhao2025llms} in the input prompt, after which the model generates a complete response from left to right. 
While this paradigm is well-suited for general open-ended generation, many real-world scenarios require models to satisfy local constraints at specific positions in the output, such as preserving a given phrase~\citep{zhou2023instruction, ifbench}, performing localized style transfer~\citep{mai2023prefix, chen2024lmstyle}, or inserting a transition or rebuttal at a designated location~\citep{lai2024style, tao2024cat}. 
For standard autoregressive (AR) LLMs, such arbitrary-position control typically requires additional constrained decoding~\citep{beurer2024guiding, banerjee2025crane}, infilling-oriented training~\citep{bavarian2022efficient, roziere2023code, ren2024empowering}, or multi-round editing processes~\citep{ge2024mart, tian2025think}, rather than being a native capability of their prefix-to-suffix generation interface.

The recent development of Diffusion Large Language Models (dLLMs) offers a natural opportunity to move beyond prefix-only interaction~\citep{nie2025large, ye2025dream, bie2025llada20}. 
Unlike autoregressive (AR) LLMs, dLLMs generate from a masked canvas through iterative bidirectional denoising, allowing fixed tokens or structured constraints to be anchored at arbitrary positions in the output sequence~\citep{arriola2025blockdiffusion, yang2025mmada}. 
This paradigm, known as In-place Prompting (IPP)~\citep{lee2025unlocking, jin2025thinking}, is important not merely as an alternative decoding format, but as a new user-facing control interface: users can specify what must happen where in the output, enabling localized editing, structured generation, precise phrase preservation, and position-specific discourse control without relying on constrained decoding or multi-round rewriting~\cite{dllmcd}.
Although recent studies have used similar anchoring mechanisms for reasoning, safety, and template infilling~\citep{jin2025thinking, horvitz2025no, zhang2025jailbreaking, lee2025unlocking, li2026diffuguard}, they mostly treat in-place spans as internal algorithmic devices rather than as an explicit instruction-following capability. 
This leaves open a central question: can dLLMs reliably execute user-specified in-place constraints while preserving natural boundary transitions and global semantic consistency?

To answer this question, we define the \textbf{I}n-place \textbf{I}nstruction \textbf{F}ollowing (\textbf{IIF}) task. 
Given a user instruction and an incomplete output canvas containing a fixed in-place constraint, IIF requires the model to generate the surrounding spans while preserving the constraint, connecting it naturally to the context, and maintaining global consistency with the user intent. 
We further construct \benchmark, a hierarchical IIF benchmark covering explanatory, knowledge-intensive, and creative generation scenarios across three progressive levels: \textbf{\ding{182} literal constraints}, \textbf{\ding{183} style constraints}, and \textbf{\ding{184} discourse-function constraints}. 
To evaluate this capability, we further design a rubric-based fine-grained local-global evaluation protocol that jointly assesses local constraint execution, boundary coherence, global consistency, and fluency, and cross-validates automatic judgments with LLM-as-a-judge, a specialized reward model, and human evaluation.

Building on \benchmark, we introduce an inference-time attention-bias probe to diagnose how dLLMs use in-place constraints during denoising. 
The probe requires no parameter updates and selectively increases attention to the IPP span. 
We find that moderate attention bias improves in-place constraint following, while overly strong bias harms generation quality, suggesting that IIF failures are partly caused by insufficient prioritization of the constraint span rather than a complete lack of linguistic capability. 
Motivated by this finding, we propose an IPP-oriented post-training framework called \textbf{G}uided \textbf{R}efinement via \textbf{A}nchor-aware \textbf{F}ocus \textbf{T}uning (\textbf{\method}). 
\method first establishes basic IIF capability through constraint-aware Supervised Fine-Tuning, which treats in-place spans as fixed conditioning variables rather than generation targets. 
It then applies constraint-aware preference optimization to improve the balance among local execution, boundary coherence, and global consistency.

The experimental results validate the effectiveness of \method framework. On 4 representative dLLMs, \method improves the average overall IIF score from 57.75 to 73.10 ($\uparrow$15.35 points), with absolute gains of 15.91 and 15.57 points on literal and discourse-function constraints, respectively. Further analysis shows that the model achieves stable improvements in local constraint execution, boundary coherence, and global consistency, while incurring only a minor impact on general generation ability. These results demonstrate that \method can effectively enhance the in-place instruction-following capability of dLLMs, providing a viable path toward coordinating localized constraints with global semantic consistency. Our contributions are summarized as follows:
\begin{itemize}[leftmargin=1em]
    \item \textbf{Task and benchmark.} We formalize the IIF task and construct \benchmark, a hierarchical benchmark covering literal, style, and discourse-function constraints.
    \item \textbf{Fine-grained evaluation.} We design a fine-grained evaluation protocol and use an attention bias probe to analyze a key bottleneck of dLLMs in following in-place constraints.
    \item \textbf{\method.} We propose \method, an IPP-oriented post-training framework that improves in-place constraint following while mitigating the conflict between local constraints and global consistency.
\end{itemize}




\begin{figure*}[t] 
    \centering
    \includegraphics[width=\textwidth]{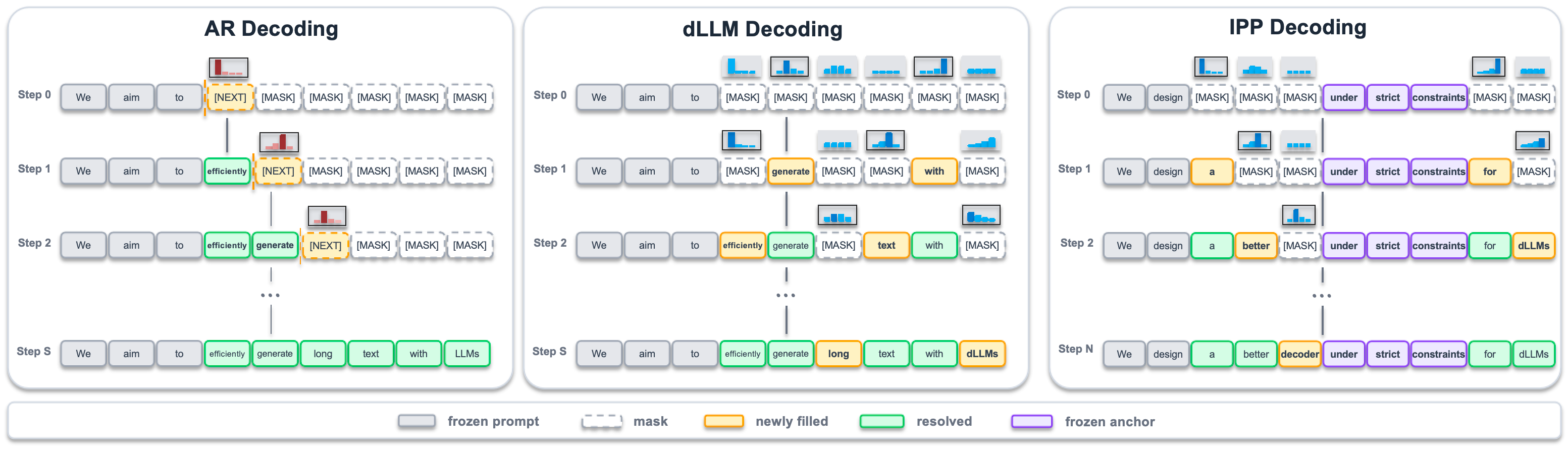} 
    \caption{Comparison of inference paradigms between standard dLLM decoding and IPP.}
    \label{fig:paradigm_comparison}
\end{figure*}
\section{Background}

\subsection{Inference Paradigm of dLLMs}
To efficiently generate long texts, dLLMs adopt a semi-autoregressive paradigm: sequentially decoding text blocks of length $K$, while internally performing $S$-step parallel denoising based on the initial state $W^{(0)} = C \oplus [\text{[MASK]}]^K$. At each step $s \in \{1, \dots, S\}$, the model predicts the distribution for the remaining masked set $\mathcal{M}_{s-1}$ given the intermediate state $W^{(s-1)}$, dynamically filtering out $\lfloor K/S \rfloor$ positions $\mathcal{I}_s$ with the highest confidence:
\begin{equation}
\resizebox{0.43\textwidth}{!}{ $
\displaystyle
\mathcal{I}_s = \operatorname{TopK}_{i \in \mathcal{M}_{s-1}} \left( \max_{v \in \mathcal{V}} p_\theta(w_i = v \mid W^{(s-1)}), \, \lfloor K/S \rfloor \right).
$ }
\end{equation}

The sequence state then undergoes an update where greedy decoding is applied to the selected positions, unselected targets are forcefully re-masked to defer their generation, and previously resolved context tokens are preserved unchanged, as formalized below:
\begin{equation}
\resizebox{0.43\textwidth}{!}{ $
\displaystyle
w_i^{(s)} = 
\begin{cases} 
\underset{v \in \mathcal{V}}{\arg\max} \, p_\theta(w_i = v \mid W^{(s-1)}), & \text{if } i \in \mathcal{I}_s, \\ 
\text{[MASK]}, & \text{if } i \in \mathcal{M}_{s-1} \setminus \mathcal{I}_s, \\ 
w_i^{(s-1)}, & \text{if } i \notin \mathcal{M}_{s-1}.
\end{cases}
$ }
\end{equation}

This predict-and-remask cycle continuously iterates ($\mathcal{M}_s = \mathcal{M}_{s-1} \setminus \mathcal{I}_s$) until the masked set is empty ($\mathcal{M}_S = \emptyset$), at which point the fully resolved text block is merged into the context $C$ to initiate the AR generation for the next block. This standard inference flow is illustrated in the middle panel of Figure~\ref{fig:paradigm_comparison}. Although dLLMs utilize parallel denoising internally, they typically follow a unidirectional trajectory where the $[\text{[MASK]}]^K$ block is appended strictly to the end of the existing context $C$. In this conventional mode, dLLM acts as a semi-autoregressive (AR) generator, extending the sequence linearly as a series of resolved blocks.

\subsection{Inference Paradigm of IPP}
The In-place Prompting (IPP) paradigm reformulates sequential generation into a constraint-aware, bidirectional infilling task within the dLLM framework. It partitions the target sequence $y$ into a prompt prefix $y_\text{prompt}$, a front generation section $y_\text{front}$ of length $L_f$, an immutable intermediate anchor $y_\text{anchor}$, and a back generation section $y_\text{back}$ of length $L_b$, denoted as $y = (y_\text{prompt}, y_\text{front}, y_\text{anchor}, y_\text{back})$. 

The inference process maps directly onto the dLLM formulation by setting the context $C = y_\text{prompt}$, the target block size $K = L_f + L_b$, and embedding the frozen anchor. The initial canvas state at denoising step $s = 0$ is formulated as:
\begin{equation}
\resizebox{0.43\textwidth}{!}{ $
\displaystyle
y^{(0)} = \left(
y_\text{prompt},
\underbrace{\text{[MASK]}, \dots, \text{[MASK]}}_{L_f \text{ tokens}}, y_\text{anchor},
\underbrace{\text{[MASK]}, \dots, \text{[MASK]}}_{L_b \text{ tokens}} \right).
$ }
\end{equation}

During the iterative refinement from step $s=1$ to $S$, the parallel predictor $p_\theta$ evaluates token distributions strictly within the remaining masked positions $\mathcal{M}_{s-1} \subseteq (y_\text{front} \cup y_\text{back})$. Crucially, $y_\text{prompt}$ and $y_\text{anchor}$ are treated as fixed context outside the dynamic masked set (i.e., $\forall i \in y_\text{anchor}, \, i \notin \mathcal{M}_{s-1}$), remaining structurally frozen as persistent conditioning variables. This forces the model to bidirectionally converge around the predefined anchor.

\section{\benchmark}
\label{sec:benchmark}
To systematically evaluate the local control capabilities of dLLMs, we propose the \textbf{I}n-place \textbf{I}nstruction \textbf{F}ollowing (\textbf{IIF}) task and construct a corresponding benchmark \textbf{\benchmark}.

\subsection{Motivation and Task Formulation}

\paragraph{Motivation.}
Unlike left-to-right AR LLMs, dLLMs support in-place prompting (IPP), allowing constraints to be inserted at arbitrary output positions. Existing benchmarks usually specify constraints upfront and focus on easily verifiable surface forms~\citep{zhou2023instruction,followbench}, leaving mid-generation intervention and its tension with global coherence largely unevaluated~\citep{zhou2023controlled,iso2024autotemplate}.

Therefore, \benchmark achieves constraint-following evaluations across three dimensions: \ding{182} It evaluates IPP capabilities within outputs, not just prefix constraints; \ding{183} it spans a three-level difficulty spectrum from literal to discourse constraints; and \ding{184} it inherently embodies and quantifies the core tension between localized constraints and global consistency (\textbf{Local-Global Tension}), making it an ideal probe for diagnosing the attention allocation mechanisms of dLLMs.
\begin{figure*}[t] 
    \centering
    \includegraphics[width=0.9\textwidth]{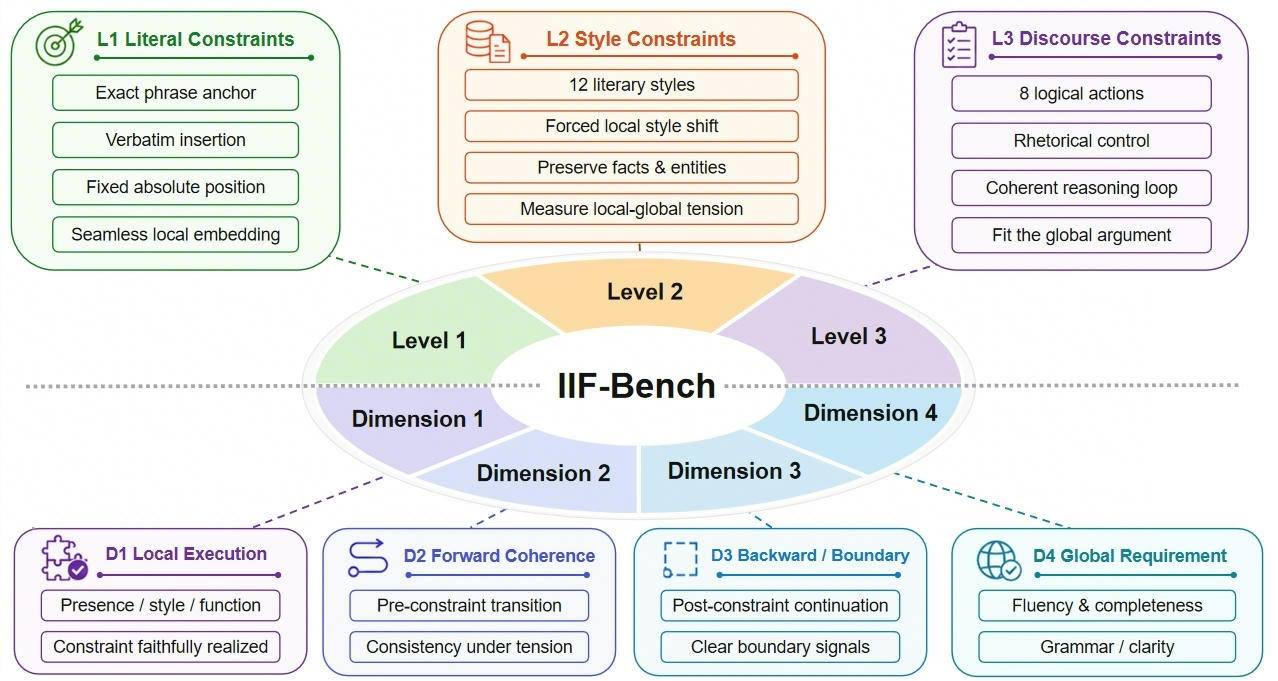} 
    \caption{Overview of the \benchmark benchmark and its evaluation protocol.}
    \label{fig:iif}
\end{figure*}

\paragraph{Task Formulation.}
Under the generalized IPP framework, we formally define the In-place Instruction Following (IIF) task. Given a user instruction represented by the prompt context $y_{\mathrm{prompt}}$ and a fixed in-place control span $y_{\mathrm{anchor}}$, wrapped by \texttt{\textless LITERAL\textgreater}, \texttt{\textless STYLE\textgreater}, or \texttt{\textless DISCOURSE\textgreater} tags, the model is required to generate the surrounding spans $y_{\mathrm{front}}$ and $y_{\mathrm{back}}$. The generated content should realize the requirement specified by $y_{\mathrm{anchor}}$ while remaining locally coherent around the anchor and globally consistent with the user instruction. Formally, letting $y_{\mathrm{gen}}=(y_{\mathrm{front}}, y_{\mathrm{back}})$, we define:

\begin{equation}
\resizebox{0.45\textwidth}{!}{$
\displaystyle
\hat{y}_{\mathrm{gen}}
=
\operatorname*{arg\,max}_{y_{\mathrm{gen}}}
p_{\theta}\!\left(y_{\mathrm{gen}} \mid y_{\mathrm{prompt}}, y_{\mathrm{anchor}}\right)
\quad
\mathrm{s.t.}
\quad
\mathcal{R}_{\ell}\!\left(\hat{y}_{\mathrm{gen}}; y_{\mathrm{anchor}}\right)=1 .
$}
\end{equation}

Here, $\mathcal{R}_{\ell}$ denotes a level-specific realization criterion. Based on the depth of control imposed by $y_\text{anchor}$ on the model's output, we divide the benchmark into 3 difficulty levels detailed in Section~\ref{dataset}.

\subsection{Dataset Introduction}
\label{dataset}

\benchmark comprises approximately 5K high-quality instances, with data sources broadly covering complex scenarios such as \datasetname{Alpaca}~\cite{alpaca}, \datasetname{LFQA}, and \datasetname{Poetry}. To comprehensively measure the model's following capabilities under varying depths of intervention, we organize the dataset into three difficulty levels:
\begin{itemize}[leftmargin=1em]
    \item \textbf{L1 Literal Constraints.} Focuses on exact replication and seamless embedding, requiring the model to use a specified exact phrase as an anchor and output it verbatim at the designated position without disrupting the syntactic structure.
    \item \textbf{L2 Style Constraints.} Focuses on the balance between forced local style transfer and global semantic tension. It requires the model to manifest one of 12 predefined literary styles within a specific paragraph. 
    \item \textbf{L3 Discourse Constraints.} Focuses on advanced rhetoric and logical deduction, requiring the model to execute 8 high-frequency logical actions within local paragraphs. It mandates that this local action form a coherent logical loop with the global core argument.
\end{itemize}

\subsection{Multi-dimensional Evaluation Protocol}

IIF requires both local constraint execution and global response quality. We therefore evaluate each output with a fine-grained local-global rubric rather than a single holistic score.
All four dimensions D1--D4 are defined per level to reflect the distinct focal points of each constraint type, while preserving a consistent evaluation structure across levels: \textbf{D1 Local Execution}, \textbf{D2 Contextual Integration}, \textbf{D3 Boundary Quality}, and \textbf{D4 Global Requirement}.

The definition of each dimension is specialized for each benchmark level.
For L1, D1 checks exact preservation and positional appropriateness of the literal anchor, and absence of paraphrastic substitution; D2 checks whether the pre-constraint text leads syntactically and semantically into the anchor; D3 checks whether the post-constraint text continues naturally from it; D4 checks instruction relevance, output completeness, and absence of verbatim input copying.
For L2, D1 checks whether the target style is realized across three independent failure modes: style onset sharpness, style depth beyond surface markers, and global style contamination; D2 checks semantic relevance of the constrained span to the preceding context, with conflict-aware adjustment described in Appendix~\ref{app:evaluation-rubrics}; D3 checks transition quality at the style tag boundary; D4 checks overall narrative integrity and instruction fulfillment.

For L3, D1 checks whether the required discourse function is correctly performed using function-specific sub-criteria (e.g., concreteness and challenge strength for \textsc{Counterexample}; conditional framing and testability for \textsc{Hypothesis}); D2 checks consistency between the discourse function segment and the global argument; D3 checks whether the function boundary is clearly demarcated; D4 checks grammatical correctness and clarity of expression within the constrained span.


To improve evaluation reliability, we use a rubric-based LLM-as-judge framework with GPT-5 as the primary evaluator and JudgeLM as an additional evaluator to enhance assessment credibility. Automatic results are reported in the main experiments, with human evaluation details provided in the appendix.

\section{Method}

Building on \benchmark, we first analyze how dLLMs use in-place constraints during iterative denoising, then improve this capability through post-training. 
We introduce an inference-time attention-bias probe to test whether failures in IIF stem from insufficient prioritization of the constraint span (Section~\ref{sec:preliminary_analysis}).
Based on the probe results, we propose an IPP-oriented post-training framework with two stages: \sft (Section~\ref{sec:sft}) and \vrpo (Section~\ref{sec:vrpo}).

\begin{figure}[tbp]
    \centering
    \includegraphics[width=\columnwidth]{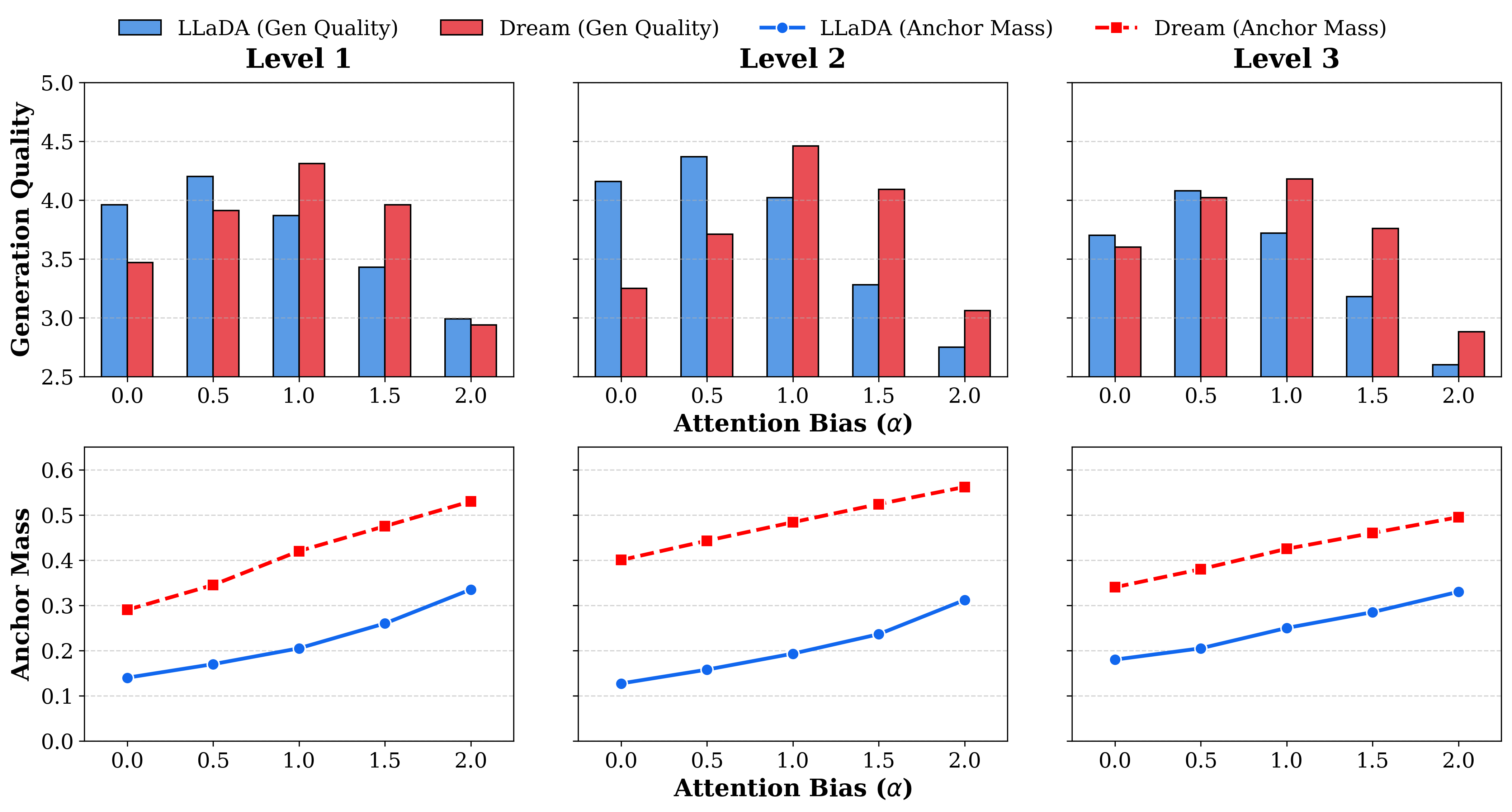}
    \caption{Comprehensive analysis of Quality and Anchor mass across different attention biases ($\alpha$).}
    \label{fig:attention_bias}
\end{figure}

\subsection{Attention Bias Probe}
\label{sec:preliminary_analysis}

To isolate the root cause of IPP failures in vanilla dLLMs, we hypothesize that these models already possess the requisite knowledge, but suffer from \textit{insufficient attention priority}. To verify this, given a set of token indices $\mathcal{S}_{\text{IPP}}$ corresponding to the IPP constraint span within the input sequence, we apply an explicit additive bias to the self-attention logits at each denoising step. For a query vector $Q_i$ and a key vector $K_j$, the modified attention weight calculation is formulated as follows:
\begin{equation}
\resizebox{0.88\linewidth}{!}{$A_{ij}=\text{Softmax}(Q_i K_j^T/\sqrt{d}+M_{ij}+\alpha\cdot\mathds{1}_{j \in \mathcal{S}_{\text{IPP}}}),$}
\label{eq:attention}
\end{equation}
where $M_{ij}$ denotes the standard attention mask, and $\alpha$ serves as a scalar hyperparameter obtained via grid search on a small-scale validation set. This intervention is implemented by registering forward pre-hooks within the dLLMs, ensuring a zero-parameter update during the inference phase. To precisely quantify the model's focus on the provided constraints during this process, we define \textbf{\textit{Anchor Mass}} as the average attention probability mass empirically allocated to the constraint tokens while generating new text.

\begin{figure}[t]
    \centering
    \includegraphics[width=\columnwidth]{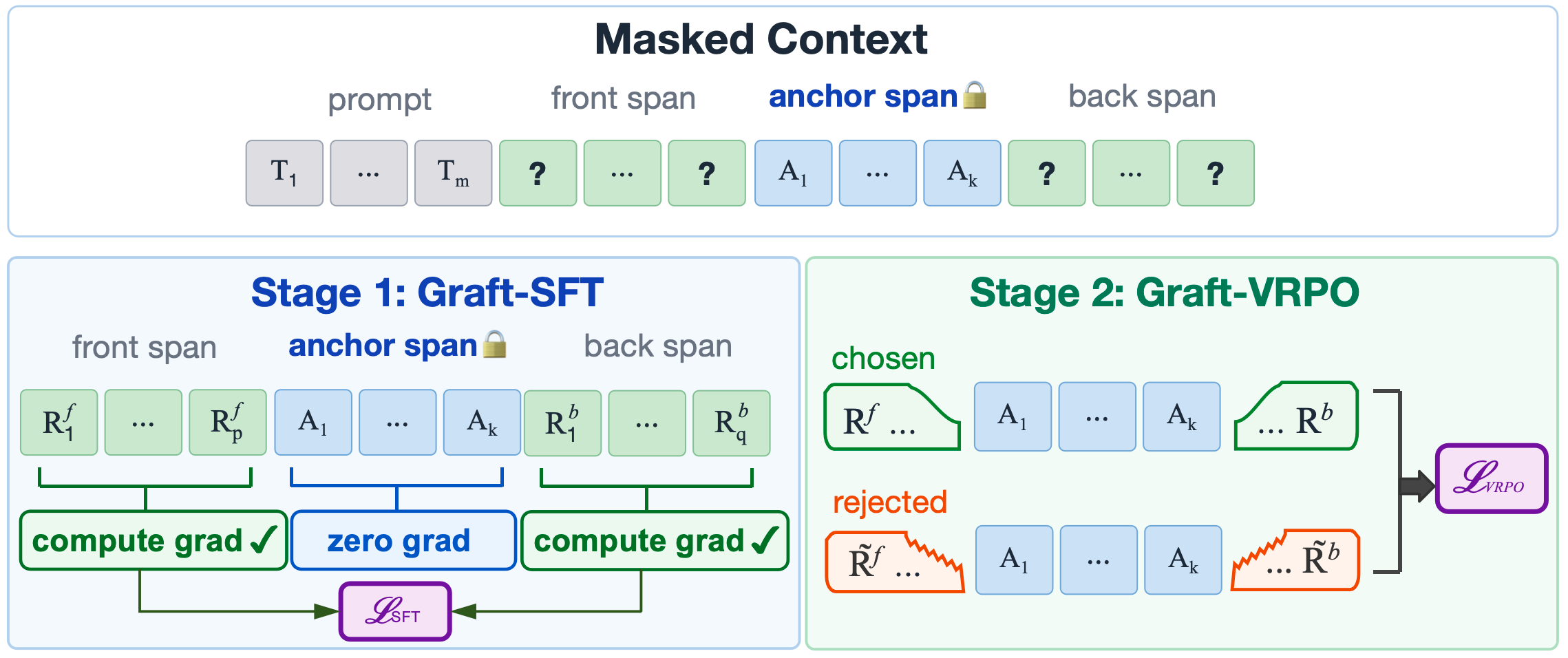}
    \caption{Overview of the proposed in-place oriented post-training framework \method.}
    \label{fig:posttraining}
\end{figure}

As shown in Figure~\ref{fig:attention_bias}, increasing $\alpha$ consistently raises the \textit{Anchor Mass} for both \llmname{LLaDA} and \llmname{Dream}, confirming that the probe effectively shifts attention toward the IPP span. 
However, IIF performance follows a non-monotonic trend: moderate bias improves constraint adherence and overall quality, whereas excessive bias overemphasizes the anchor and degrades fluency and contextual fit. 
These results suggest that vanilla dLLMs are not inherently unable to follow in-place constraints; rather, they often fail to assign the constraint span an appropriate level of priority during denoising. 
This motivates our post-training framework, which aims to internalize such constraint-aware behavior into the model parameters.

\subsection{In-place Oriented Post-Training}
\label{sec:post_training}

To internalize the inference-time attention bias discovered in our preliminary analysis into the model parameters, we propose \method, an In-place Oriented Post-Training framework. This framework consists of two progressive stages: \sft and \vrpo. The key design of this post-training framework is that its training objective is optimized exclusively for the IPP paradigm: namely, regarding the user-inserted anchor as a fixed contextual constraint, instead of a target to be generated.

\subsubsection{\sft}
\label{sec:sft}
To equip the base dLLM with the fundamental ability to execute in-place instructions and maintain boundary smoothness, we first perform \sft. Unlike standard SFT that calculates the cross-entropy loss over the entire sequence, our objective avoids predicting the constraint itself.

Given the target token sequence
\[
y = [y_\text{prompt}, y_\text{front}, y_\text{anchor}, y_\text{back}],
\]
the anchor $y_\text{anchor}$ acts as a fixed conditioning variable. Calculating the loss over the anchor tokens would misguide the model into memorizing the specific constraint rather than learning the bidirectional contextual transition. To address this, we apply a localized masking strategy to the loss function. Let $\mathcal{S}_\text{gen}$ denote the set of token indices corresponding exclusively to the generated spans $y_\text{front}$ and $y_\text{back}$. The \sft loss is formulated as:
\begin{equation}
\label{eq:sft}
\begin{aligned}
\mathcal{L}_\text{\sft} &= \mathbb{E}_{s, W^{(s)}} \left[ - \sum_{i \in \mathcal{S}_\text{gen}} \ell_\theta^{(i)} \right], \\
\ell_\theta^{(i)} &= \log p_\theta\!\left(y_i \mid W^{(s)}, y_\text{prompt}, y_\text{anchor}\right),
\end{aligned}
\end{equation}
where $W^{(s)}$ represents the intermediate masked state at diffusion step $s$. By zeroing out the loss gradients on the anchor positions where $i \notin \mathcal{S}_\text{gen}$, the model is forced to dedicate its parameter updates exclusively to harmonizing the surrounding context with the given local constraints.

\subsubsection{\vrpo}
\label{sec:vrpo}
While \sft establishes basic instruction-following behavior, navigating the delicate trade-off between local constraint execution and global semantic consistency remains challenging, especially under extreme stylistic or discourse conflicts such as L2 and L3 constraints in \benchmark. To further align the model's behavior, we introduce \vrpo to further align the model toward responses that better balance local anchor adherence and global consistency. We construct preference pairs consisting of a chosen response $y_w$ and a rejected response $y_l$, where $y_w$ better balances the local anchor adherence and global consistency. Similarly to the \sft phase, the preference optimization is strictly evaluated on the generated spans $\mathcal{S}_\text{gen}$, treating $y_\text{anchor}$ as a grounded condition. The modified objective is defined as:
\begin{equation}
\mathcal{L}_\text{\vrpo} = -\mathbb{E}_{x, y_w, y_l} \Bigl[ \log \sigma \Bigl( \Delta_w - \Delta_l \Bigr) \Bigr],
\label{eq:vrpo}
\end{equation}
where
\begin{equation}
\begin{aligned}
\Delta_w &= \beta \log \frac{\pi_\theta(y_w \mid x, y_\text{anchor})}{\pi_\text{ref}(y_w \mid x, y_\text{anchor})}, \\
\Delta_l &= \beta \log \frac{\pi_\theta(y_l \mid x, y_\text{anchor})}{\pi_\text{ref}(y_l \mid x, y_\text{anchor})}.
\end{aligned}
\end{equation}

Here $x$ denotes the combined input context including $y_\text{prompt}$.

\section{Experiments}

\subsection{Setup}
We conduct \method training and evaluation on \benchmark{} with four diffusion LLM backbones: \llmname{LLaDA-1.5}, \llmname{LLaDA-8B}, \llmname{Dream-7B}, and \llmname{MMaDA}.

Baselines include their vanilla diffusion counterparts and three prompt-based AR instruction-tuned models: \llmname{GPT-4o}, \llmname{Llama-3.1-8B-Instruct}, and \llmname{Qwen3-8B-Instruct}. 
Since AR models cannot natively denoise masked spans around a fixed in-place anchor, we convert each IIF instance into a natural-language prompt specifying the anchor content, constraint type, and intended position, and ask the model to generate a complete response satisfying this positional constraint. 
Thus, these AR baselines test whether left-to-right instruction following can approximate IIF without native IPP-style denoising. 

We further compare with our aligned variants after \sft and the full SFT-to-DPO pipeline. 
Following prior LLM-as-a-judge protocols, we use \llmname{GPT-5} and \llmname{JudgeLM}~\citep{zhu2025judgelm} as evaluators and report scores averaged over three runs; detailed settings are provided in Appendix~\ref{app:exp-details}.


\subsection{Overall Performance}
\begin{figure}[t]
    \centering
    \includegraphics[width=\columnwidth]{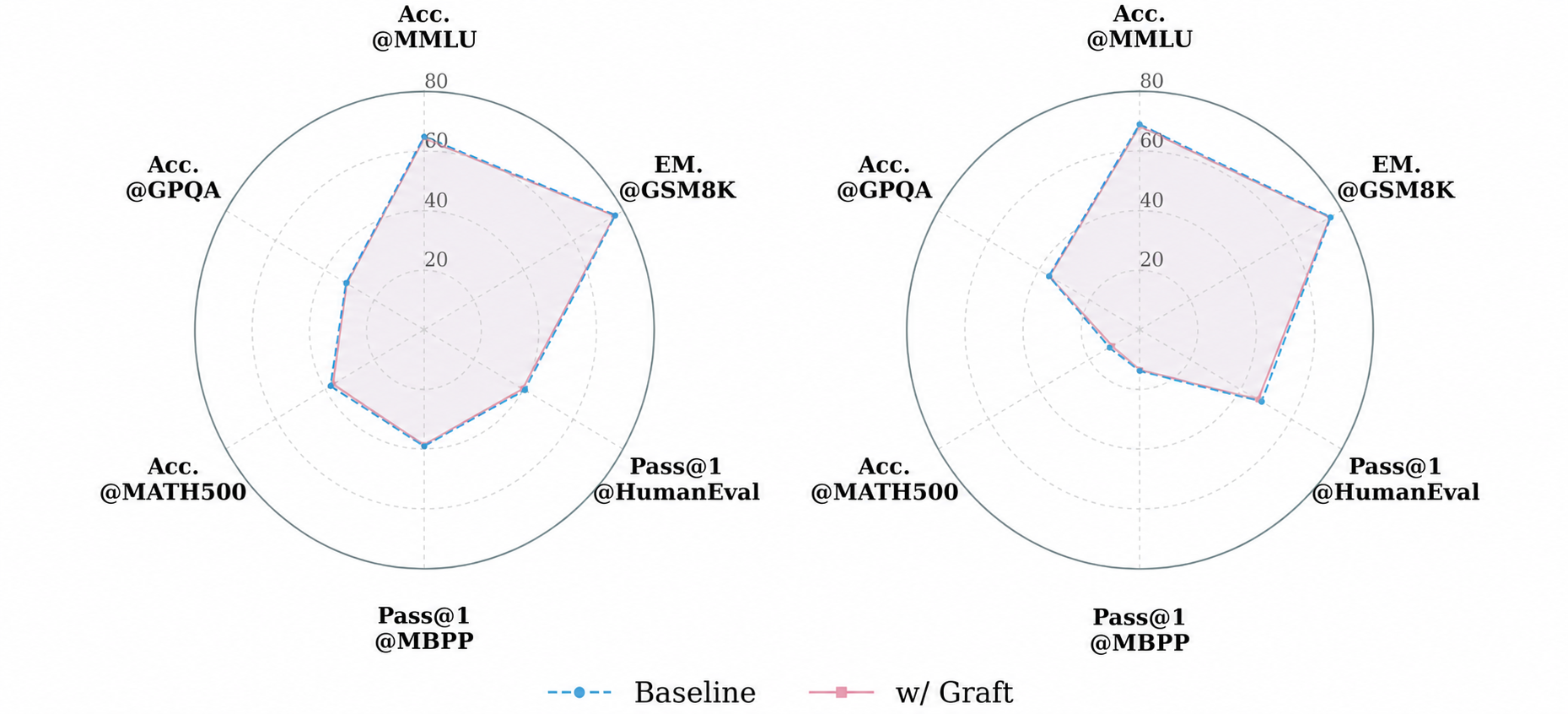}
    \caption{Performance comparison of LLaDA (left) and Dream (right) across multiple general benchmarks.}
    \label{fig:general}
\end{figure}

As shown in Table~\ref{tab:per_level_evaluation}, \method-aligned dLLMs consistently outperform their vanilla counterparts across all constraint levels, yielding an average absolute improvement of over 15 points in overall score. Although GPT-4o slightly leads at the highest constraint complexity ($L_3$), our models heavily dominate at the $L_1$ and $L_2$ levels. Ablation results confirm that \method-SFT establishes the foundational capability, while full \method alignment yields a universal boost. Furthermore, Fig. \ref{fig:general} shows that \method-aligned models maintain robust general capabilities, indicating that \method enhances in-place constraint following at negligible cost.

\begin{table*}[t]
\centering
\scriptsize
\vspace{2mm}
\setlength{\tabcolsep}{3pt}
\resizebox{\textwidth}{!}{%
\begin{tabular}{clcccccccc}
\toprule
\multirow{2}{*}{\rotatebox[origin=c]{90}{}}
& \multirow{2}{*}{\textbf{Model \& Stage}} 
& \multicolumn{4}{c}{\textbf{GPT-5 Judge}} 
& \multicolumn{4}{c}{\textbf{JudgeLM}} \\
\cmidrule(lr){3-6} \cmidrule(lr){7-10}
& & \textbf{L1} & \textbf{L2} & \textbf{L3} & \textbf{Overall}
  & \textbf{L1} & \textbf{L2} & \textbf{L3} & \textbf{Overall} \\
\midrule

\multirow{3}{*}{\rotatebox[origin=c]{90}{\textbf{AR}}}
& GPT-4o & 63.74 & 67.18 & \textbf{70.42} & 67.11 & 63.39 & 64.98 & \textbf{65.57} & 64.65 \\
& Llama-3.1-8B-Instruct & 58.74 & 62.36 & 65.18 & 62.09 & 57.79 & 59.74 & 61.58 & 59.70 \\
& Qwen3-8B-Instruct & 61.42 & 65.83 & 67.56 & 64.94 & 60.82 & 62.98 & 63.49 & 62.43 \\

\midrule

\multirow{16}{*}{\rotatebox[origin=c]{90}{\textbf{Aligned dLLMs (Ours)}}}
& \modelrow{LLaDA-1.5} \\
\cmidrule{2-10}
& \quad $\hookrightarrow$ \vanilla{Vanilla}
& \vanilla{67.42} & \vanilla{64.51} & \vanilla{56.59} & \vanilla{62.84}
& \vanilla{66.32} & \vanilla{61.81} & \vanilla{53.49} & \vanilla{60.54} \\
& \quad $\hookrightarrow$ + only \method-SFT 
& 75.86\gain{8.44} & 71.42\gain{6.91} & 64.70\gain{8.11} & 70.66\gain{7.82}
& 75.41\gain{9.09} & 69.50\gain{7.69} & 59.60\gain{6.11} & 68.17\gain{7.63} \\
& \quad $\hookrightarrow$ + \method 
& \underline{78.93}\gain{11.51} & \textbf{75.64}\gain{11.13} & 68.09\gain{11.50} & \underline{74.22}\gain{11.38}
& \underline{78.28}\gain{11.96} & \underline{73.32}\gain{11.51} & 63.89\gain{10.40} & \underline{71.83}\gain{11.29} \\

\cmidrule{2-10}
& \modelrow{LLaDA-8B} \\
\cmidrule{2-10}
& \quad $\hookrightarrow$ \vanilla{Vanilla}
& \vanilla{62.18} & \vanilla{58.91} & \vanilla{51.38} & \vanilla{57.49}
& \vanilla{61.38} & \vanilla{56.40} & \vanilla{47.52} & \vanilla{55.10} \\
& \quad $\hookrightarrow$ + only \method-SFT 
& 76.14\gain{13.96} & 70.42\gain{11.51} & 65.12\gain{13.74} & 70.56\gain{13.07}
& 75.84\gain{14.46} & 68.37\gain{11.97} & 59.92\gain{12.40} & 68.04\gain{12.94} \\
& \quad $\hookrightarrow$ + \method 
& \textbf{80.17}\gain{17.99} & \underline{75.24}\gain{16.33} & \underline{69.53}\gain{18.15} & \textbf{74.98}\gain{17.49}
& \textbf{79.97}\gain{18.59} & \textbf{73.39}\gain{16.99} & \underline{64.08}\gain{16.56} & \textbf{72.48}\gain{17.38} \\

\cmidrule{2-10}
& \modelrow{Dream-7B} \\
\cmidrule{2-10}
& \quad $\hookrightarrow$ \vanilla{Vanilla}
& \vanilla{59.83} & \vanilla{56.94} & \vanilla{48.65} & \vanilla{55.14}
& \vanilla{58.98} & \vanilla{54.54} & \vanilla{44.65} & \vanilla{52.72} \\
& \quad $\hookrightarrow$ + only \method-SFT 
& 73.84\gain{14.01} & 68.51\gain{11.57} & 62.76\gain{14.11} & 68.37\gain{13.23}
& 73.29\gain{14.31} & 66.06\gain{11.52} & 58.51\gain{13.86} & 65.95\gain{13.23} \\
& \quad $\hookrightarrow$ + \method 
& 77.84\gain{18.01} & 72.41\gain{15.47} & 66.59\gain{17.94} & 72.28\gain{17.14}
& 77.44\gain{18.46} & 70.31\gain{15.77} & 61.94\gain{17.29} & 69.90\gain{17.18} \\

\cmidrule{2-10}
& \modelrow{MMaDA} \\
\cmidrule{2-10}
& \quad $\hookrightarrow$ \vanilla{Vanilla}
& \vanilla{60.27} & \vanilla{56.12} & \vanilla{50.21} & \vanilla{55.53}
& \vanilla{59.22} & \vanilla{53.62} & \vanilla{46.71} & \vanilla{53.18} \\
& \quad $\hookrightarrow$ + only \method-SFT 
& 72.18\gain{11.91} & 66.92\gain{10.80} & 61.12\gain{10.91} & 66.74\gain{11.21}
& 71.48\gain{12.26} & 64.62\gain{11.00} & 56.82\gain{10.11} & 64.31\gain{11.13} \\
& \quad $\hookrightarrow$ + \method 
& 76.42\gain{16.15} & 71.43\gain{15.31} & 64.91\gain{14.70} & 70.92\gain{15.39}
& 75.87\gain{16.65} & 69.11\gain{15.49} & 60.61\gain{13.90} & 68.53\gain{15.35} \\

\bottomrule
\end{tabular}
}
\caption{Comprehensive evaluation on \benchmark{}. We compare our aligned diffusion LLMs against vanilla diffusion baselines and advanced AR models. \textbf{Bold} indicates the best performance for each metric, and \underline{underline} indicates the second-best performance. Green subscripts denote absolute gains over the corresponding vanilla model. Metrics are averaged over 3 independent runs.}
\label{tab:per_level_evaluation}
\end{table*}

\subsection{Attention Diagnosis}
\begin{figure}[t]
    \centering
    \includegraphics[width=\columnwidth]{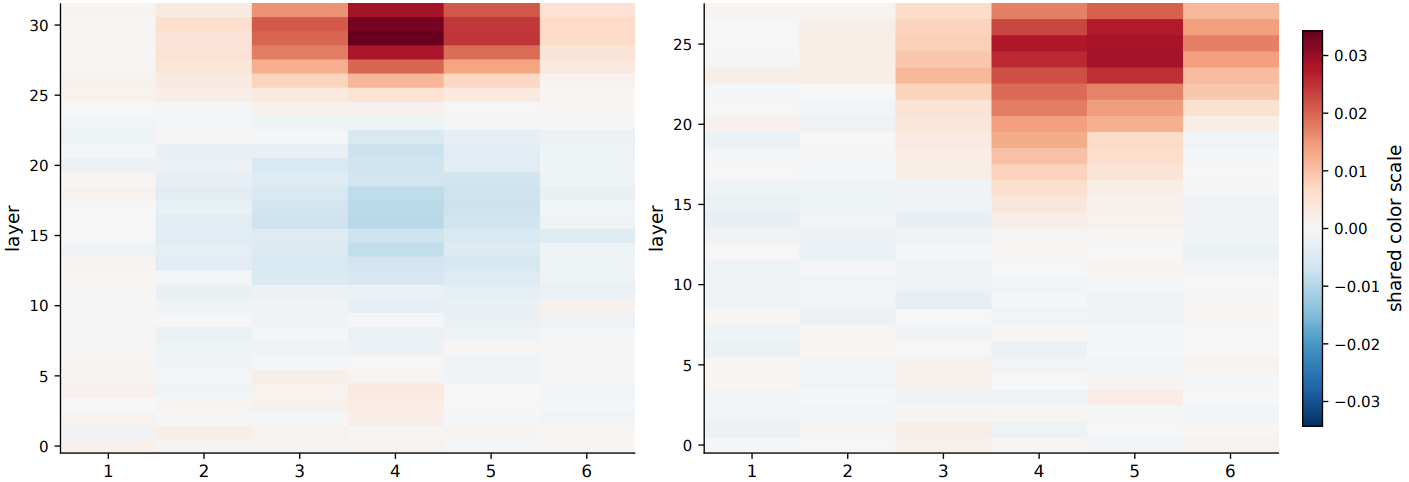}
\caption{Layer-step difference maps of the anchor-probe score for \llmname{LLaDA} (left) and \llmname{Dream} (right).}
    \label{fig:anchor_probe}
\end{figure}

To understand how \method affects dLLM denoising, we visualize layer-step difference maps of the anchor-probe score in Fig.~\ref{fig:anchor_probe}, where $\Delta=\text{\method}-\text{Vanilla}$. The x-axis denotes six uniformly sampled probe checkpoints along the denoising trajectory. Positive values indicate stronger alignment with the in-place anchor after \method, while negative values indicate weaker anchor-related similarity or attention. For \llmname{LLaDA}, the largest positive change appears in upper layers, while middle layers show slight decreases. For \llmname{Dream}, the positive change is smoother and extends from upper layers to middle layers. These results suggest that \method does not overwrite the whole pretrained denoising process. Instead, it selectively adjusts task-relevant upper-layer representations to better use the in-place constraint.

\subsection{Ablation Study}
As shown in Tables~\ref{tab:freq-penalty-ablation} 
and~\ref{tab:suppress-tag-ablation}, we conduct 
ablation studies on frequency penalty and structural 
tag suppression. The frequency penalty lowers the 
logits of already-generated tokens to reduce 
repetition, while structural tag suppression blocks 
tag-related tokens from being generated. A moderate 
frequency penalty consistently yields optimal 
performance across both vanilla and \method-aligned 
models. Notably, \method-aligned models exhibit a 
markedly flatter trajectory in the low-penalty range 
($fp \in [0.0, 0.3]$), indicating that anti-repetition 
behavior has been internalized through post-training 
and no longer relies heavily on test-time logit 
intervention. The comparable degradation observed 
in both model variants under excessive penalty 
($fp \geq 0.7$) reflects task-structural damage, 
where necessary literal phrases and discourse 
connectors are over-suppressed regardless of 
alignment stage. For structural tag suppression, 
test-time blocking improves vanilla baselines by 
preventing metadata leaks. After \method alignment, 
this gain narrows and converges across architectures, 
indicating that post-training establishes native 
formatting boundaries and reduces reliance on 
test-time logit interventions.

\begin{table}[t]
\centering
\small
\vspace{1mm}
\resizebox{\columnwidth}{!}{
\begin{tabular}{lccccc}
\toprule
Model \& Stage & fp=0.0 & fp=0.1 & fp=0.3 & fp=0.7 & fp=1.0 \\
\midrule
LLaDA-8B (Vanilla) & 57.49 & 57.88 & 58.32 & 57.65 & 55.42 \\
LLaDA-8B (+\method) & 74.98 & 75.12 & 75.23 & 74.15 & 72.31 \\
\midrule
Dream-7B (Vanilla) & 55.14 & 55.42 & 55.68 & 53.81 & 51.24 \\
Dream-7B (+\method) & 72.28 & 72.39 & 72.46 & 70.82 & 68.54 \\
\bottomrule
\end{tabular}%
} 
\caption{Ablation of frequency penalty $fp$ for vanilla and aligned \method models.}
\label{tab:freq-penalty-ablation}
\end{table}

\begin{table}[t]
\centering
\small
\vspace{1mm}
\resizebox{\columnwidth}{!}{
\begin{tabular}{lccccc}
\toprule
Model \& Stage & Tag & L1 & L2 & L3 & Overall \\
\midrule
\multirow{2}{*}{LLaDA-8B (Vanilla)} & Off & 62.18 & 58.91 & 51.38 & 57.49 \\
 & On & 62.35 & 60.41 & 52.14 & 58.30 \\
\midrule
\multirow{2}{*}{LLaDA-8B (+\method)} & Off & 80.17 & 75.24 & 69.53 & 74.98 \\
 & On & 80.24 & 75.68 & 69.71 & 75.21 \\
\midrule
\multirow{2}{*}{Dream-7B (Vanilla)} & Off & 59.83 & 56.94 & 48.65 & 55.14 \\
 & On & 59.94 & 58.15 & 49.10 & 55.73 \\
\midrule
\multirow{2}{*}{Dream-7B (+\method)} & Off & 77.84 & 72.41 & 66.59 & 72.28 \\
 & On & 77.92 & 72.84 & 66.82 & 72.53 \\
\bottomrule
\end{tabular}%
} 
\caption{Ablation of the anchor tag word suppression across different levels with GPT-5 Score.}
\label{tab:suppress-tag-ablation}
\end{table}

\section{Related Work}

\paragraph{Diffusion Language Models.}
Diffusion language models adapt denoising diffusion to text generation. Early approaches mainly explore continuous denoising in embedding space or task-specific sequence generation~\citep{li2022diffusionlm,gong2023diffuseq,he2023diffusionbert}, while discrete diffusion models introduce token-level corruption processes such as absorbing masks~\citep{austin2021d3pm,lou2023discrete}. 
Recent masked diffusion language models have substantially improved the scalability and generation quality of this paradigm. LLaDA~\citep{nie2025large} trains a large-scale masked diffusion language model from scratch and demonstrates competitive instruction-following and in-context learning ability; Dream~\citep{ye2025dream} adapts autoregressive LLMs into diffusion models and improves the quality--efficiency trade-off; LLaDA-1.5~\citep{zhu2025llada} investigates post-training and preference optimization for diffusion LMs; and LLaDA-2.X series~\citep{bie2025llada20, bie2026llada2} further scale up masked diffusion models to 100B parameters.
The paradigm has also been extended to multimodal generation through MMaDA~\citep{yang2025mmada} and LLaDA-V~\citep{you2025llada}, and recent work studies inference acceleration and block-wise generation for more efficient dLLM deployment~\citep{liu2025dllmcache,arriola2025blockdiffusion,cheng2025sdar,wu2025fastdllmv2}. 
Compared with autoregressive LMs, these models perform iterative masked-token refinement with bidirectional context, which makes them naturally suitable for infilling and arbitrary-position control. This property provides the modeling basis for our study of in-place instruction following.

\paragraph{In-place Prompting and Mid-sequence Control.}
Recent work explores dLLMs beyond prefix-only prompting. 
ICE inserts chain-of-thought prompts into masked positions with confidence-aware early exit~\citep{jin2025thinking}, while Template Infilling uses structural anchors as global blueprints for masked segment generation~\citep{lee2025unlocking}. 
Reasoning-as-infilling fills structured reasoning and answer templates for posterior reasoning and early exits~\citep{horvitz2025no}, and DIP dynamically inserts in-context examples during denoising~\citep{li2026dip}. 
Related safety studies show that the same bidirectional and parallel mechanisms create diffusion-specific attack surfaces~\citep{wen2025devil,zhang2025jailbreaking,li2026diffuguard}. 
Unlike these works, we study in-place instruction following as a user-facing mechanism for localized constraint satisfaction under global coherence.

\section{Conclusion}

This paper introduces and formalizes the In-Place Instruction Following task and constructs a benchmark spanning three difficulty levels. Through an attention-bias probe, we find evidence that insufficient prioritization of the constraint span is one source of failure in vanilla dLLMs during parallel denoising. To address this, we propose \method, a post-training framework combining \sft and \vrpo. Additionally, a fine-grained evaluation reveals a persistent trade-off between local and global consistency in style constraint scenarios.

\section*{Limitations}

While \method improves in-place instruction following in dLLMs, this work has several limitations. First, \benchmark{} mainly focuses on text-only in-place instruction following, and its coverage of multilingual, multimodal, and interactive editing scenarios remains limited. Second, although the proposed evaluation protocol combines automatic, reward-model-based, and human judgments, rubric-based evaluation may still be affected by evaluator bias, especially for style and discourse-function constraints. Third, \method is evaluated on four representative dLLMs, but its scalability to larger diffusion models and its robustness under more complex or conflicting anchors require further investigation. Future work may extend IIF to broader domains and develop more interpretable mechanisms for local--global constraint coordination.

\bibliography{main}

\begin{thebibliography}{50}
\providecommand{\natexlab}[1]{#1}

\bibitem[{Arriola et~al.(2025)Arriola, Gokaslan, Chiu, Yang, Qi, Han, Sahoo, and Kuleshov}]{arriola2025blockdiffusion}
Marianne Arriola, Aaron Gokaslan, Justin~T. Chiu, Zhihan Yang, Zhixuan Qi, Jiaqi Han, Subham~Sekhar Sahoo, and Volodymyr Kuleshov. 2025.
\newblock Block diffusion: Interpolating between autoregressive and diffusion language models.
\newblock \emph{arXiv preprint arXiv:2503.09573}.

\bibitem[{Austin et~al.(2021)Austin, Johnson, Ho, Tarlow, and van~den Berg}]{austin2021d3pm}
Jacob Austin, Daniel~D. Johnson, Jonathan Ho, Daniel Tarlow, and Rianne van~den Berg. 2021.
\newblock Structured denoising diffusion models in discrete state-spaces.
\newblock In \emph{Advances in Neural Information Processing Systems}.

\bibitem[{Banerjee et~al.(2025)Banerjee, Suresh, Ugare, Misailovic, and Singh}]{banerjee2025crane}
Debangshu Banerjee, Tarun Suresh, Shubham Ugare, Sasa Misailovic, and Gagandeep Singh. 2025.
\newblock Crane: Reasoning with constrained llm generation.
\newblock \emph{arXiv preprint arXiv:2502.09061}.

\bibitem[{Bavarian et~al.(2022)Bavarian, Jun, Tezak, Schulman, McLeavey, Tworek, and Chen}]{bavarian2022efficient}
Mohammad Bavarian, Heewoo Jun, Nikolas Tezak, John Schulman, Christine McLeavey, Jerry Tworek, and Mark Chen. 2022.
\newblock Efficient training of language models to fill in the middle.
\newblock \emph{arXiv preprint arXiv:2207.14255}.

\bibitem[{Beurer-Kellner et~al.(2024)Beurer-Kellner, Fischer, and Vechev}]{beurer2024guiding}
Luca Beurer-Kellner, Marc Fischer, and Martin Vechev. 2024.
\newblock Guiding llms the right way: Fast, non-invasive constrained generation.
\newblock \emph{arXiv preprint arXiv:2403.06988}.

\bibitem[{Bie et~al.(2026)Bie, Cao, Cao, Chen, Chen, Chen, Du, Feng, Feng, Gong et~al.}]{bie2026llada2}
Tiwei Bie, Maosong Cao, Xiang Cao, Bingsen Chen, Fuyuan Chen, Kun Chen, Lun Du, Daozhuo Feng, Haibo Feng, Mingliang Gong, and 1 others. 2026.
\newblock Llada2. 1: Speeding up text diffusion via token editing.
\newblock \emph{arXiv preprint arXiv:2602.08676}.

\bibitem[{Bie et~al.(2025)Bie, Cao, Chen, Du, Gong, Gong, Gu, Hu, Huang, Lan, Li, Li, Li, Li, Liu, Liu, Lu, Lu, Ma, Tan, Wei, Wen, Xing, Zhang, Zhao, Zheng, Zhou, Zhou, Zhou, Zhu, and Zhuang}]{bie2025llada20}
Tiwei Bie, Maosong Cao, Kun Chen, Lun Du, Mingliang Gong, Zhuochen Gong, Yanmei Gu, Jiaqi Hu, Zenan Huang, Zhenzhong Lan, Chengxi Li, Chongxuan Li, Jianguo Li, Zehuan Li, Huabin Liu, Ling Liu, Guoshan Lu, Xiaocheng Lu, Yuxin Ma, and 12 others. 2025.
\newblock Llada2.0: Scaling up diffusion language models to 100b.
\newblock \emph{arXiv preprint arXiv:2512.15745}.

\bibitem[{Brown et~al.(2020)Brown, Mann, Ryder, Subbiah, Kaplan, Dhariwal, Neelakantan, Shyam, Sastry, Askell, Agarwal, Herbert-Voss, Krueger, Henighan, Child, Ramesh, Ziegler, Wu, Winter, Hesse, Chen, Sigler, Litwin, Gray, Chess, Clark, Berner, McCandlish, Radford, Sutskever, and Amodei}]{gpt3}
Tom~B. Brown, Benjamin Mann, Nick Ryder, Melanie Subbiah, Jared Kaplan, Prafulla Dhariwal, Arvind Neelakantan, Pranav Shyam, Girish Sastry, Amanda Askell, Sandhini Agarwal, Ariel Herbert-Voss, Gretchen Krueger, Tom Henighan, Rewon Child, Aditya Ramesh, Daniel~M. Ziegler, Jeffrey Wu, Clemens Winter, and 12 others. 2020.
\newblock \href {https://arxiv.org/abs/2005.14165} {Language models are few-shot learners}.
\newblock \emph{Preprint}, arXiv:2005.14165.

\bibitem[{Chen(2024)}]{chen2024lmstyle}
Jianlin Chen. 2024.
\newblock Lmstyle benchmark: Evaluating text style transfer for chatbots.
\newblock \emph{arXiv preprint arXiv:2403.08943}.

\bibitem[{Cheng et~al.(2025)Cheng, Bian, Liu, Jiang, Liu, Zhang, Wang, Guo, Chen, Qi, and Zhou}]{cheng2025sdar}
Shuang Cheng, Yihan Bian, Dawei Liu, Yuhua Jiang, Yihao Liu, Linfeng Zhang, Wenhai Wang, Qipeng Guo, Kai Chen, Biqing Qi, and Bowen Zhou. 2025.
\newblock Sdar: A synergistic diffusion-autoregression paradigm for scalable sequence generation.
\newblock \emph{arXiv preprint arXiv:2510.06303}.

\bibitem[{Ge et~al.(2024)Ge, Zhou, Hou, Khabsa, Wang, Wang, Han, and Mao}]{ge2024mart}
Suyu Ge, Chunting Zhou, Rui Hou, Madian Khabsa, Yi-Chia Wang, Qifan Wang, Jiawei Han, and Yuning Mao. 2024.
\newblock Mart: Improving llm safety with multi-round automatic red-teaming.
\newblock In \emph{Proceedings of the 2024 Conference of the North American Chapter of the Association for Computational Linguistics: Human Language Technologies (Volume 1: Long Papers)}, pages 1927--1937.

\bibitem[{Gong et~al.(2023)Gong, Li, Feng, Wu, and Kong}]{gong2023diffuseq}
Shansan Gong, Mukai Li, Jiangtao Feng, Zhiyong Wu, and Lingpeng Kong. 2023.
\newblock Diffuseq: Sequence to sequence text generation with diffusion models.
\newblock In \emph{International Conference on Learning Representations}.

\bibitem[{He et~al.(2023)He, Sun, Wang, Huang, and Qiu}]{he2023diffusionbert}
Zhengfu He, Tianxiang Sun, Kuanning Wang, Xuanjing Huang, and Xipeng Qiu. 2023.
\newblock Diffusionbert: Improving generative masked language models with diffusion models.
\newblock In \emph{Proceedings of the 61st Annual Meeting of the Association for Computational Linguistics}.

\bibitem[{Horvitz et~al.(2025)Horvitz, Singhal, Zou, Domingo-Enrich, Yu, Ranganath, and McKeown}]{horvitz2025no}
Zachary Horvitz, Raghav Singhal, Hao Zou, Carles Domingo-Enrich, Zhou Yu, Rajesh Ranganath, and Kathleen McKeown. 2025.
\newblock No compute left behind: Rethinking reasoning and sampling with masked diffusion models.
\newblock \emph{arXiv preprint arXiv:2510.19990}.

\bibitem[{Iso(2024)}]{iso2024autotemplate}
Hayate Iso. 2024.
\newblock Autotemplate: A simple recipe for lexically constrained text generation.
\newblock In \emph{Proceedings of the 17th International Natural Language Generation Conference}, pages 1--12.

\bibitem[{Jiang et~al.(2024)Jiang, Wang, Zeng, Zhong, Li, Mi, Shang, Jiang, Liu, and Wang}]{followbench}
Yuxin Jiang, Yufei Wang, Xingshan Zeng, Wanjun Zhong, Liangyou Li, Fei Mi, Lifeng Shang, Xin Jiang, Qun Liu, and Wei Wang. 2024.
\newblock Followbench: A multi-level fine-grained constraints following benchmark for large language models.
\newblock In \emph{Proceedings of the 62nd Annual Meeting of the Association for Computational Linguistics (Volume 1: Long Papers)}, pages 4667--4688.

\bibitem[{Jin et~al.(2025)Jin, Wang, Gao, Wen, Qi, Liu, and Zhang}]{jin2025thinking}
Xiangqi Jin, Yuxuan Wang, Yifeng Gao, Zichen Wen, Biqing Qi, Dongrui Liu, and Linfeng Zhang. 2025.
\newblock Thinking inside the mask: In-place prompting in diffusion llms.
\newblock \emph{arXiv preprint arXiv:2508.10736}.

\bibitem[{Lai et~al.(2024)Lai, Hangya, and Fraser}]{lai2024style}
Wen Lai, Viktor Hangya, and Alexander Fraser. 2024.
\newblock Style-specific neurons for steering llms in text style transfer.
\newblock In \emph{Proceedings of the 2024 Conference on Empirical Methods in Natural Language Processing}, pages 13427--13443.

\bibitem[{Lee et~al.(2025)Lee, Kim, and Kwak}]{lee2025unlocking}
Junhoo Lee, Seungyeon Kim, and Nojun Kwak. 2025.
\newblock Unlocking the potential of diffusion language models through template infilling.
\newblock \emph{arXiv preprint arXiv:2510.13870}.

\bibitem[{Li et~al.(2024)Li, Zhou, Sun, Zhang, Zhao, and Liu}]{li2024dissecting}
Junlong Li, Fan Zhou, Shichao Sun, Yikai Zhang, Hai Zhao, and Pengfei Liu. 2024.
\newblock Dissecting human and llm preferences.
\newblock In \emph{Proceedings of the 62nd Annual Meeting of the Association for Computational Linguistics (Volume 1: Long Papers)}, pages 1790--1811.

\bibitem[{Li et~al.(2022)Li, Thickstun, Gulrajani, Liang, and Hashimoto}]{li2022diffusionlm}
Xiang~Lisa Li, John Thickstun, Ishaan Gulrajani, Percy Liang, and Tatsunori~B. Hashimoto. 2022.
\newblock Diffusion-lm improves controllable text generation.
\newblock In \emph{Advances in Neural Information Processing Systems}.

\bibitem[{Li et~al.(2026{\natexlab{a}})Li, Meng, Wang, and Chen}]{li2026dip}
Yang Li, Han Meng, Chenan Wang, and Haipeng Chen. 2026{\natexlab{a}}.
\newblock Dip: Dynamic in-context planner for diffusion language models.
\newblock \emph{arXiv preprint arXiv:2601.03199}.

\bibitem[{Li et~al.(2026{\natexlab{b}})Li, Nie, Zhou, Liu, Zhang, Cheng, Wen, Wang, Guo, and Zhang}]{li2026diffuguard}
Zherui Li, Zheng Nie, Zhenhong Zhou, Yue Liu, Yitong Zhang, Yu~Cheng, Qingsong Wen, Kun Wang, Yufei Guo, and Jiaheng Zhang. 2026{\natexlab{b}}.
\newblock Diffuguard: How intrinsic safety is lost and found in diffusion large language models.
\newblock In \emph{International Conference on Learning Representations}.

\bibitem[{Liu et~al.(2024)Liu, Liu, Fiannaca, Koo, Dixon, Terry, and Cai}]{liu2024we}
Michael~Xieyang Liu, Frederick Liu, Alexander~J Fiannaca, Terry Koo, Lucas Dixon, Michael Terry, and Carrie~J Cai. 2024.
\newblock " we need structured output": Towards user-centered constraints on large language model output.
\newblock In \emph{Extended Abstracts of the CHI Conference on Human Factors in Computing Systems}, pages 1--9.

\bibitem[{Liu et~al.(2025)Liu, Yang, Zhang, Chen, Zou, Wei, Wang, and Zhang}]{liu2025dllmcache}
Zhiyuan Liu, Yicun Yang, Yaojie Zhang, Junjie Chen, Chang Zou, Qingyuan Wei, Shaobo Wang, and Linfeng Zhang. 2025.
\newblock dllm-cache: Accelerating diffusion large language models with adaptive caching.
\newblock \emph{arXiv preprint arXiv:2506.06295}.

\bibitem[{Lou et~al.(2023)Lou, Meng, and Ermon}]{lou2023discrete}
Aaron Lou, Chenlin Meng, and Stefano Ermon. 2023.
\newblock Discrete diffusion modeling by estimating the ratios of the data distribution.
\newblock \emph{arXiv preprint arXiv:2310.16834}.

\bibitem[{Lou et~al.(2024)Lou, Zhang, and Yin}]{if-survey}
Renze Lou, Kai Zhang, and Wenpeng Yin. 2024.
\newblock \href {https://doi.org/10.1162/coli_a_00523} {Large language model instruction following: A survey of progresses and challenges}.
\newblock \emph{Computational Linguistics}, 50(3):1053--1095.

\bibitem[{Mai et~al.(2023)Mai, Jiang, and Deng}]{mai2023prefix}
Huiyu Mai, Wenhao Jiang, and Zhi-Hong Deng. 2023.
\newblock Prefix-tuning based unsupervised text style transfer.
\newblock In \emph{Findings of the Association for Computational Linguistics: EMNLP 2023}, pages 14847--14856.

\bibitem[{Nie et~al.(2025)Nie, Zhu, You, Zhang, Ou, Hu, Zhou, Lin, Wen, and Li}]{nie2025large}
Shen Nie, Fengqi Zhu, Zebin You, Xiaolu Zhang, Jingyang Ou, Jun Hu, Jun Zhou, Yankai Lin, Ji-Rong Wen, and Chongxuan Li. 2025.
\newblock Large language diffusion models.
\newblock \emph{arXiv preprint arXiv:2502.09992}.

\bibitem[{Pyatkin et~al.(2026)Pyatkin, Malik, Graf, Ivison, Huang, Dasigi, Lambert, and Hajishirzi}]{ifbench}
Valentina Pyatkin, Saumya Malik, Victoria Graf, Hamish Ivison, Shengyi Huang, Pradeep Dasigi, Nathan Lambert, and Hanna Hajishirzi. 2026.
\newblock Generalizing verifiable instruction following.
\newblock \emph{Advances in Neural Information Processing Systems}, 38.

\bibitem[{Radford et~al.(2019)Radford, Wu, Child, Luan, Amodei, Sutskever et~al.}]{gpt2}
Alec Radford, Jeffrey Wu, Rewon Child, David Luan, Dario Amodei, Ilya Sutskever, and 1 others. 2019.
\newblock Language models are unsupervised multitask learners.
\newblock \emph{OpenAI blog}, 1(8):9.

\bibitem[{Ren et~al.(2024)Ren, Zhan, Wu, and Li}]{ren2024empowering}
Houxing Ren, Mingjie Zhan, Zhongyuan Wu, and Hongsheng Li. 2024.
\newblock Empowering character-level text infilling by eliminating sub-tokens.
\newblock In \emph{Proceedings of the 62nd Annual Meeting of the Association for Computational Linguistics (Volume 1: Long Papers)}, pages 3253--3267.

\bibitem[{Roziere et~al.(2023)Roziere, Gehring, Gloeckle, Sootla, Gat, Tan, Adi, Liu, Sauvestre, Remez et~al.}]{roziere2023code}
Baptiste Roziere, Jonas Gehring, Fabian Gloeckle, Sten Sootla, Itai Gat, Xiaoqing~Ellen Tan, Yossi Adi, Jingyu Liu, Romain Sauvestre, Tal Remez, and 1 others. 2023.
\newblock Code llama: Open foundation models for code.
\newblock \emph{arXiv preprint arXiv:2308.12950}.

\bibitem[{Son et~al.(2024)Son, Baek, Nam, Jeong, and Kim}]{multitask}
Guijin Son, SangWon Baek, Sangdae Nam, Ilgyun Jeong, and Seungone Kim. 2024.
\newblock Multi-task inference: Can large language models follow multiple instructions at once?
\newblock In \emph{Proceedings of the 62nd Annual Meeting of the Association for Computational Linguistics (Volume 1: Long Papers)}, pages 5606--5627.

\bibitem[{Tao et~al.(2024)Tao, Xi, Li, Tang, and Xu}]{tao2024cat}
Zhen Tao, Dinghao Xi, Zhiyu Li, Liumin Tang, and Wei Xu. 2024.
\newblock Cat-llm: Style-enhanced large language models with text style definition for chinese article-style transfer.
\newblock \emph{arXiv preprint arXiv:2401.05707}.

\bibitem[{Taori et~al.(2023)Taori, Gulrajani, Zhang, Dubois, Li, Guestrin, Liang, and Hashimoto}]{alpaca}
Rohan Taori, Ishaan Gulrajani, Tianyi Zhang, Yann Dubois, Xuechen Li, Carlos Guestrin, Percy Liang, and Tatsunori~B. Hashimoto. 2023.
\newblock Stanford alpaca: An instruction-following llama model.
\newblock \url{https://github.com/tatsu-lab/stanford_alpaca}.

\bibitem[{Tian et~al.(2025)Tian, Zhao, Wang, Chen, Ji, Peng, Zhao, and Li}]{tian2025think}
Xiaoyu Tian, Sitong Zhao, Haotian Wang, Shuaiting Chen, Yunjie Ji, Yiping Peng, Han Zhao, and Xiangang Li. 2025.
\newblock Think twice: Enhancing llm reasoning by scaling multi-round test-time thinking.
\newblock \emph{arXiv preprint arXiv:2503.19855}.

\bibitem[{Wen et~al.(2025)Wen, Qu, Chen, Lu, Liu, Liu, Wu, Yang, Jin, Xu et~al.}]{wen2025devil}
Zichen Wen, Jiashu Qu, Zhaorun Chen, Xiaoya Lu, Dongrui Liu, Zhiyuan Liu, Ruixi Wu, Yicun Yang, Xiangqi Jin, Haoyun Xu, and 1 others. 2025.
\newblock The devil behind the mask: An emergent safety vulnerability of diffusion llms.
\newblock \emph{arXiv preprint arXiv:2507.11097}.

\bibitem[{Wu et~al.(2025)Wu, Zhang, Xue, Diao, Fu, Liu, Molchanov, Luo, Han, and Xie}]{wu2025fastdllmv2}
Chengyue Wu, Hao Zhang, Shuchen Xue, Shizhe Diao, Yonggan Fu, Zhijian Liu, Pavlo Molchanov, Ping Luo, Song Han, and Enze Xie. 2025.
\newblock Fast-dllm v2: Efficient block-diffusion llm.
\newblock \emph{arXiv preprint arXiv:2509.26328}.

\bibitem[{Yang et~al.(2025)Yang, Tian, Li, Zhang, Shen, Tong, and Wang}]{yang2025mmada}
Ling Yang, Ye~Tian, Bowen Li, Xinchen Zhang, Ke~Shen, Yunhai Tong, and Mengdi Wang. 2025.
\newblock Mmada: Multimodal large diffusion language models.
\newblock \emph{arXiv preprint arXiv:2505.15809}.

\bibitem[{Ye et~al.(2025)Ye, Xie, Zheng, Gao, Wu, Jiang, Li, and Kong}]{ye2025dream}
Jiacheng Ye, Zhihui Xie, Lin Zheng, Jiahui Gao, Zirui Wu, Xin Jiang, Zhenguo Li, and Lingpeng Kong. 2025.
\newblock Dream 7b: Diffusion large language models.
\newblock \emph{arXiv preprint arXiv:2508.15487}.

\bibitem[{You et~al.(2025)You, Nie, Zhang, Hu, Zhou, Lu, Wen, and Li}]{you2025llada}
Zebin You, Shen Nie, Xiaolu Zhang, Jun Hu, Jun Zhou, Zhiwu Lu, Ji-Rong Wen, and Chongxuan Li. 2025.
\newblock Llada-v: Large language diffusion models with visual instruction tuning.
\newblock \emph{arXiv preprint arXiv:2505.16933}.

\bibitem[{Zhang et~al.(2026)Zhang, Li, Liu, Li, Jia, Li, and Li}]{dllmcd}
Yitong Zhang, Yongmin Li, Yuetong Liu, Jia Li, Xiaoran Jia, Zherui Li, and Ge~Li. 2026.
\newblock Lookahead-then-verify: Reliable constrained decoding for diffusion llms under context-free grammars.
\newblock \emph{arXiv preprint arXiv:2602.00612}.

\bibitem[{Zhang et~al.(2025)Zhang, Xie, Zhou, Li, Chen, Wang, and Guo}]{zhang2025jailbreaking}
Yuanhe Zhang, Fangzhou Xie, Zhenhong Zhou, Zherui Li, Hao Chen, Kun Wang, and Yufei Guo. 2025.
\newblock Jailbreaking large language diffusion models: Revealing hidden safety flaws in diffusion-based text generation.
\newblock \emph{arXiv preprint arXiv:2507.19227}.

\bibitem[{Zhao et~al.(2025{\natexlab{a}})Zhao, Hong, Liu, Hazarika, and Lin}]{zhao2025llms}
Siyan Zhao, Mingyi Hong, Yang Liu, Devamanyu Hazarika, and Kaixiang Lin. 2025{\natexlab{a}}.
\newblock Do llms recognize your preferences? evaluating personalized preference following in llms.
\newblock \emph{arXiv preprint arXiv:2502.09597}.

\bibitem[{Zhao et~al.(2025{\natexlab{b}})Zhao, Zhou, Li, Tang, Wang, Hou, Min, Zhang, Zhang, Dong, Du, Yang, Chen, Chen, Jiang, Ren, Li, Tang, Liu, Liu, Nie, and Wen}]{llm-survey}
Wayne~Xin Zhao, Kun Zhou, Junyi Li, Tianyi Tang, Xiaolei Wang, Yupeng Hou, Yingqian Min, Beichen Zhang, Junjie Zhang, Zican Dong, Yifan Du, Chen Yang, Yushuo Chen, Zhipeng Chen, Jinhao Jiang, Ruiyang Ren, Yifan Li, Xinyu Tang, Zikang Liu, and 3 others. 2025{\natexlab{b}}.
\newblock \href {https://arxiv.org/abs/2303.18223} {A survey of large language models}.
\newblock \emph{Preprint}, arXiv:2303.18223.

\bibitem[{Zhou et~al.(2023{\natexlab{a}})Zhou, Lu, Mishra, Brahma, Basu, Luan, Zhou, and Hou}]{zhou2023instruction}
Jeffrey Zhou, Tianjian Lu, Swaroop Mishra, Siddhartha Brahma, Sujoy Basu, Yi~Luan, Denny Zhou, and Le~Hou. 2023{\natexlab{a}}.
\newblock Instruction-following evaluation for large language models.
\newblock \emph{arXiv preprint arXiv:2311.07911}.

\bibitem[{Zhou et~al.(2023{\natexlab{b}})Zhou, Jiang, Wilcox, Cotterell, and Sachan}]{zhou2023controlled}
Wangchunshu Zhou, Yuchen~Eleanor Jiang, Ethan Wilcox, Ryan Cotterell, and Mrinmaya Sachan. 2023{\natexlab{b}}.
\newblock Controlled text generation with natural language instructions.
\newblock In \emph{International Conference on Machine Learning}, pages 42602--42613. PMLR.

\bibitem[{Zhu et~al.(2025{\natexlab{a}})Zhu, Wang, Nie, Zhang, Wu, Hu, Zhou, Chen, Lin, Wen et~al.}]{zhu2025llada}
Fengqi Zhu, Rongzhen Wang, Shen Nie, Xiaolu Zhang, Chunwei Wu, Jun Hu, Jun Zhou, Jianfei Chen, Yankai Lin, Ji-Rong Wen, and 1 others. 2025{\natexlab{a}}.
\newblock Llada 1.5: Variance-reduced preference optimization for large language diffusion models.
\newblock \emph{arXiv preprint arXiv:2505.19223}.

\bibitem[{Zhu et~al.(2025{\natexlab{b}})Zhu, Wang, and Wang}]{zhu2025judgelm}
Lianghui Zhu, Xinggang Wang, and Xinlong Wang. 2025{\natexlab{b}}.
\newblock Judgelm: Fine-tuned large language models are scalable judges.
\newblock In \emph{International Conference on Learning Representations}, volume 2025, pages 51257--51296.

\end{thebibliography}

\appendix

\newpage

\section{LLM Usage Statement}
We utilized Large Language Models to refine and polish our original manuscript. Specifically, its use was focused on improving grammar, clarity, conciseness, and word choice. It is important to note that the model was employed solely as a writing aid and did not contribute to the generation of any new content or ideas.

\section{Experimental Details}
\label{app:exp-details}

\paragraph{Implementation details.}
All training experiments are conducted on NVIDIA A800 GPUs. 
We apply the same \method pipeline to each diffusion backbone, including LLaDA-1.5, LLaDA-8B, Dream-7B, and MMaDA. 
During training, the inserted constraint is fixed in the output canvas, and the model is optimized to generate the surrounding masked spans. 
At inference time, all diffusion baselines and \method-aligned variants follow the same IPP decoding format for fair comparison.

\paragraph{Benchmark construction.}
We construct the IIF benchmark by converting existing instruction--response pairs into
partially constrained continuation tasks. Each example consists of a user instruction and
an assistant-side sequence in which a constraint tag is inserted between a prefix
(\textit{front text}) and a target continuation (\textit{back text}). The benchmark contains
three levels of increasing constraint abstraction. Level 1 evaluates literal adherence:
the continuation must preserve either an exact sentence or a set of verbatim keyword
phrases sampled from the original answer. Level 2 evaluates semantic style control:
poetry-style responses are split into a prefix and a continuation, and the continuation
is rewritten under one of twelve target style constraints while preserving the core
content. Level 3 evaluates discourse-level control: long-form QA responses are split into
a prefix and continuation, and the continuation is rewritten to perform one of eight
specified discourse functions, such as giving a counterexample, forming an analogy, or
posing rhetorical questions. We use Alpaca and LFQA for literal constraints, Poetry
Instructions for style constraints, and LFQA for discourse-function constraints. Tables~\ref{tab:l2-style-inventory} and~\ref{tab:l3-discourse-inventory} summarize the constraint inventories used for Level 2 and Level 3, respectively. Table~\ref{tab:iif-benchmark-statistics} reports the number of examples and constraint-length statistics for each level of the IIF benchmark.
\begin{table}[t]
\centering
\small
\resizebox{\columnwidth}{!}{%
\begin{tabular}{lrrrr}
\toprule
\textbf{Level} & \textbf{\#Examples} & \textbf{Avg. constraint length} & \textbf{Min.} & \textbf{Max.} \\
\midrule
Level 1 Literal   & 1573 & 129.5 & 54  & 395 \\
Level 2 Semantic  & 1628 & 114.5 & 107 & 134 \\
Level 3 Discourse & 1836 & 96.2  & 82  & 106 \\
\midrule
Total             & 5037 & 112.5 & 54  & 395 \\
\bottomrule
\end{tabular}%
}
\caption{Statistics of constraint lengths across the three levels of the IIF benchmark.}
\label{tab:iif-benchmark-statistics}
\end{table}

\paragraph{Training and preference data construction.}
We use the constructed IIF instances as the data source for both \sft and \vrpo.
For each instance, the constraint-satisfying rewritten response produced by our construction pipeline is used as the target response for \sft.
For \vrpo, we form preference pairs by treating the constraint-satisfying rewritten response as the chosen response and its corresponding unconstrained or pre-rewrite response as the rejected response.
This construction is applied consistently across L1 literal, L2 style, and L3 discourse-function constraints.
Training and evaluation are conducted on disjoint splits of the constructed IIF instances.

\paragraph{Baselines.}
We compare \method with both diffusion-based and autoregressive language model baselines. 
For diffusion LLMs, all vanilla models are evaluated under the same IPP format as \method, where the inserted constraint is fixed in the output canvas and the model generates the surrounding masked spans. 
For autoregressive models, which generate text from left to right, we prompt them to produce a complete response satisfying the same in-place constraint specification.

\begin{itemize}
    \item \textbf{LLaDA-1.5 and LLaDA-8B.}
    LLaDA represents the masked-diffusion generation paradigm, where masked response tokens are iteratively recovered through parallel denoising. 
    We include LLaDA-8B and LLaDA-1.5 as representative models from this diffusion LLM family, evaluated without \method-specific adaptation.

    \item \textbf{Dream-7B.}
    Dream-7B is another discrete diffusion LLM that supports iterative parallel refinement and infilling-style generation. 
    It allows us to evaluate \method on a diffusion backbone developed independently of LLaDA.

    \item \textbf{MMaDA.}
    MMaDA is a unified multimodal diffusion model that formulates text generation through masked-token prediction under a shared diffusion architecture. 
    Its inclusion broadens the evaluation to a diffusion backbone with a different architectural origin from both LLaDA and Dream.

    \item \textbf{GPT-4o.} 
    GPT-4o is included as a strong closed-source autoregressive baseline. 
    Since it does not natively perform masked-span denoising around a fixed output constraint, its performance mainly reflects whether a high-capability autoregressive model can satisfy the in-place constraint through prompt understanding.

    \item \textbf{Llama-3.1-8B-Instruct.} 
    Llama-3.1-8B-Instruct is used as an open-weight instruction-tuned autoregressive baseline at a scale comparable to the 7B--8B diffusion backbones. 
    This comparison helps distinguish the benefit of diffusion-style in-place generation from standard autoregressive instruction following.

    \item \textbf{Qwen3-8B-Instruct.} 
    Qwen3-8B-Instruct is included as another open-weight 8B-scale autoregressive instruction-tuned baseline. 
    Together with Llama-3.1-8B-Instruct, it provides a diverse open-model comparison for evaluating constraint-following performance under the same benchmark.
\end{itemize}

\paragraph{Inference hyperparameters.}
For diffusion LLMs, the generation length is instance-specific. 
Given each benchmark example, we construct the output canvas as 
\texttt{[prompt | MASK $\times L_f$ | anchor | MASK $\times L_b$]}, 
where the anchor span contains the fixed inserted constraint. 
The effective generation length is therefore $L_f + L_b$, while the full sequence length is capped at 1024 tokens. 
If the constructed sequence exceeds this cap, $L_f$ and $L_b$ are proportionally reduced. 
The inference hyperparameters are summarized in Table~\ref{tab:inference_params}.

\begin{table}[t]
\centering
\small
\resizebox{\columnwidth}{!}{
\begin{tabular}{lcccccccc}
\toprule
\textbf{Model} 
& \textbf{Steps} 
& \textbf{Gen. Length} 
& \textbf{Block Length} 
& \textbf{Temp.} 
& \textbf{CFG Scale} 
& \textbf{Top-$p$} 
& \textbf{Alg.} 
& \textbf{Max Length} \\
\midrule
LLaDA-1.5 & 128 & $L_f + L_b$ & 32 & 0.5 & 0.0 & --   & --           & 1024 \\
LLaDA-8B  & 128 & $L_f + L_b$ & 32 & 0.5 & 0.0 & --   & --           & 1024 \\
MMaDA     & 128 & $L_f + L_b$ & 32 & 0.5 & 0.0 & --   & --           & 1024 \\
Dream-7B  & 128 & $L_f + L_b$ & -- & 0.7 & --  & 0.95 & maskgit-plus & 1024 \\
\bottomrule
\end{tabular}
}
\caption{Inference hyperparameters for diffusion LLMs. The generation length is instance-specific and equals the number of masked tokens surrounding the fixed constraint span.}
\label{tab:inference_params}
\end{table}

\begin{table*}[t]
\centering
\small
\setlength{\tabcolsep}{5pt}
\renewcommand{\arraystretch}{1.08}
\begin{tabular}{
  P{0.7cm}
  L{12.8cm}
}
\toprule
\textbf{ID} & \textbf{Target Style Description} \\
\midrule
1
& haunting Gothic, heavy with dread, ruin, and shadowed decay \\

2
& solemn and elegiac, like a funeral lament for something lost \\

3
& wistful and nostalgic, like a faded letter reread years later \\

4
& stoic and pastoral, the quiet voice of a farmer watching seasons turn \\

5
& mystical and reverent, in the cadence of sacred liturgy or prayer \\

6
& austere and minimalist, in the spirit of a Zen haiku---few words, vast silence \\

7
& tender and amorous, in the warm voice of a Neruda love sonnet \\

8
& serene and epiphanic, a quiet moment of sudden inner light \\

9
& jubilant and celebratory, in the expansive voice of a Whitman ode \\

10
& joyful pastoral, bright with sunlight, birdsong, and dancing streams \\

11
& whimsical and playful, in the nonsense-verse tradition of Lear or Carroll \\

12
& surreal and dreamlike, where images dissolve into one another without logical bridges \\
\bottomrule
\end{tabular}
\caption{Level-2 semantic style constraint inventory.
Each L2 instance samples one target style from this inventory and requires the continuation span to instantiate the assigned style while preserving the core content.}
\label{tab:l2-style-inventory}
\end{table*}

\begin{table*}[t]
\centering
\small
\setlength{\tabcolsep}{5pt}
\renewcommand{\arraystretch}{1.08}
\begin{tabular}{
  P{0.7cm}
  L{3.3cm}
  L{9.3cm}
}
\toprule
\textbf{ID} & \textbf{Function} & \textbf{Core Requirement} \\
\midrule
1
& \textsc{Counterexample}
& Provide a concrete case that contradicts, challenges, or complicates the implication of the preceding context. \\

2
& \textsc{Hypothesis}
& Introduce a speculative ``Suppose / Imagine / What if'' scenario that extends or stress-tests the preceding claim. \\

3
& \textsc{Summary}
& Compress the preceding context into a broader takeaway or general principle without introducing new arguments. \\

4
& \textsc{Transition}
& Briefly acknowledge the preceding point and then pivot to a contrasting or complementary perspective. \\

5
& \textsc{Analogy}
& Explain the preceding point through a concrete cross-domain comparison. \\

6
& \textsc{Rebuttal}
& Identify and contest an implicit assumption in the preceding context, then offer a counterclaim with support. \\

7
& \textsc{Parallelism}
& Continue the discourse using three or more structurally parallel clauses or sentences. \\

8
& \textsc{Rhetorical Question}
& End with one or more unanswered questions that deepen or reinforce the preceding point. \\
\bottomrule
\end{tabular}
\caption{Level-3 discourse function inventory.
Each L3 instance samples one discourse function and requires the continuation span to perform the specified rhetorical or logical role with respect to the preceding context.}
\label{tab:l3-discourse-inventory}
\end{table*}

\section{Extended Experimental Analysis}
\label{app:exp}

\paragraph{Ablation on Anchor Positions}
As shown in Table~\ref{tab:position-ablation}, we analyze the spatial generalization of \method by partitioning single anchor placement into Early (10\%--30\%), Middle (40\%--60\%), and Late (70\%--90\%) segments. Both LLaDA-8B and Dream-7B exhibit a clear middle anchor advantage, where the overall scores peak at the centrally located configuration. This stems from the bidirectional mechanism of In-place Prompting, as a centralized anchor provides a balanced semantic bridge to restrict both flanking spans. Conversely, extreme positioning leaves one side disproportionately long, causing the task to partially degenerate into unconstrained sequential generation. 

Interestingly, the two architectures diverge at the Late position for stylized constraints ($L_2$). LLaDA shows a performance upturn because its block-wise semi-autoregressive decoding allows longer preceding context blocks to undergo sufficient refinement iterations for style manifestation. In contrast, Dream experiences degradation because its continuous diffusion hook mechanism suffers from attenuated control signals over an expanded attention radius, hindering global stylistic consistency.

\begin{table}[htbp]
\centering
\small
\resizebox{\columnwidth}{!}{%
\begin{tabular}{llcccc}
\toprule
\textbf{Model \& Stage} & \textbf{Position} & \textbf{L1} & \textbf{L2} & \textbf{L3} & \textbf{Overall} \\
\midrule
\multirow{3}{*}{LLaDA-8B (+\method)}
 & Early  & 79.62 & 74.21 & 68.95 & 74.26 \\
 & Middle & 80.54 & 75.12 & 70.43 & 75.36 \\
 & Late   & 79.88 & 75.81 & 68.74 & 74.81 \\
\midrule
\multirow{3}{*}{Dream-7B (+\method)}
 & Early  & 77.12 & 71.15 & 66.18 & 71.48 \\
 & Middle & 78.11 & 73.24 & 66.92 & 72.76 \\
 & Late   & 77.93 & 72.08 & 65.74 & 71.92 \\
\bottomrule
\end{tabular}%
}
\caption{Ablation of single anchor positions.}
\label{tab:position-ablation}
\end{table}

\paragraph{Scaling of Multiple Anchors}
Table~\ref{tab:multi-anchor-ablation} illustrates the impact of scaling the number of uniformly distributed anchors from one to three. Both models follow a sub-linear saturation degradation trajectory, where the transition from one to two anchors induces a severe performance fracture, especially in discourse-function constraints ($L_3$). This sharp decline highlights a critical generalization boundary for \method post-training. Since the training objective focuses on single-anchor instances, the internalized attention priorities are calibrated for a single focal point. 

When multiple anchors are introduced, the limited attention resources become fragmented across competing constraints. The model consequently struggles to maintain local execution for each individual anchor while simultaneously ensuring global coherence across fragmented generation segments. Literal constraints ($L_1$) remain the most resilient due to the lack of combinatorial logic or style collisions, suggesting that while \method effectively internalizes single-point attention bias, multi-focal coordination remains a structural hurdle.

\begin{table}[htbp]
\centering
\small
\resizebox{\columnwidth}{!}{%
\begin{tabular}{llcccc}
\toprule
\textbf{Model \& Stage} & \textbf{\# Anchors} & \textbf{L1} & \textbf{L2} & \textbf{L3} & \textbf{Overall} \\
\midrule
\multirow{3}{*}{LLaDA-8B (+\method)}
 & 1 & 80.23 & 75.11 & 69.64 & 75.03 \\
 & 2 & 77.86 & 70.58 & 63.91 & 70.78 \\
 & 3 & 76.04 & 66.73 & 58.42 & 67.06 \\
\midrule
\multirow{3}{*}{Dream-7B (+\method)}
 & 1 & 77.69 & 72.36 & 66.71 & 72.25 \\
 & 2 & 74.92 & 67.18 & 60.46 & 67.52 \\
 & 3 & 72.81 & 63.27 & 55.04 & 63.71 \\
\bottomrule
\end{tabular}%
}
\caption{Ablation of anchor quantity scaling with uniform distribution.}
\label{tab:multi-anchor-ablation}
\end{table}

\begin{table}[t]
\centering
\scriptsize
\setlength{\tabcolsep}{3pt}
\renewcommand{\arraystretch}{0.95}

\resizebox{\columnwidth}{!}{%
\begin{tabular}{llc}
\toprule
\textbf{Dim} & \textbf{Description} & \textbf{Max pts} \\
\midrule
C1 & Register distance (formality gap between global and local style) & 35 \\
C2 & Emotional valence conflict (opposing affective directions)        & 35 \\
C3 & Structural incompatibility (local style conflicts with global genre form)     & 30 \\
\midrule
   & \textbf{Total ConflictScore}                                      & \textbf{100} \\
\bottomrule
\end{tabular}%
}
\caption{Dimensions used to compute the L2 ConflictScore.}
\label{tab:iif-conflict}
\end{table}

Table~\ref{tab:per_dim_evaluation_human} breaks down the scores by dimension
for the double-blind human evaluation subset
introduced in Section~\ref{sec:benchmark}. Scores are decomposed into
the four rubric dimensions D1--D4 defined in
Appendix~\ref{app:evaluation-rubrics} and complement the holistic
scores reported in Table~\ref{tab:per_level_evaluation}. 

\section{Human Evaluation Details}
\label{app:human-eval-details}

To empirically validate the reliability of our automated evaluation pipeline, we conduct a double-blind human study on a randomly sampled subset of \benchmark{}. Table~\ref{tab:per_dim_evaluation_human} provides the fine-grained, per-dimension breakdown across all three difficulty levels, extending the holistic analysis provided in the main text. 

\begin{table*}[t]
\centering
\scriptsize
\setlength{\tabcolsep}{3.5pt}
\renewcommand{\arraystretch}{1.05}
\resizebox{\textwidth}{!}{%
\begin{tabular}{llcccc|cccc|cccc|c}
\toprule
\multirow{2}{*}{\textbf{Model Archetype}}
& \multirow{2}{*}{\textbf{Model \& Stage}}
& \multicolumn{4}{c|}{\textbf{L1} (40/20/20/20)}
& \multicolumn{4}{c|}{\textbf{L2} (40/25$^*$/20/15)}
& \multicolumn{4}{c|}{\textbf{L3} (40/30/20/10)}
& \multirow{2}{*}{\textbf{Overall}} \\
\cmidrule(lr){3-6} \cmidrule(lr){7-10} \cmidrule(lr){11-14}
& & \textbf{D1} & \textbf{D2} & \textbf{D3} & \textbf{D4}
  & \textbf{D1} & \textbf{D2}$^*$ & \textbf{D3} & \textbf{D4}
  & \textbf{D1} & \textbf{D2} & \textbf{D3} & \textbf{D4} & \\
\midrule
\multirow{4}{*}{\shortstack[l]{\textbf{Diffusion LLMs}\\\textbf{Vanilla Baselines}}}
& LLaDA-1.5                 & 23.33 & 15.00 & 14.67 & 14.00 & 22.67 & 18.55 & 13.00 & 10.33 & 19.00 & 18.67 & 11.00 & 7.00 & 62.41 \\
& LLaDA-8B                  & 20.67 & 14.00 & 13.33 & 13.00 & 20.67 & 17.43 & 11.33 & 9.33  & 16.67 & 17.00 & 10.00 & 6.67 & 56.70 \\
& Dream-7B                  & 19.33 & 13.67 & 13.00 & 13.00 & 19.33 & 17.35 & 10.67 & 9.33  & 15.33 & 16.00 & 9.33  & 7.00 & 54.45 \\
& MMaDA                     & 19.67 & 13.67 & 13.33 & 13.33 & 19.00 & 16.98 & 11.00 & 9.00  & 16.00 & 16.67 & 9.67  & 7.00 & 55.11 \\
\midrule
\multirow{3}{*}{\shortstack[l]{\textbf{Autoregressive}\\\textbf{Baselines}}}
& GPT-4o                    & 22.67 & 14.33 & 13.67 & 13.33 & 27.33 & 18.87 & 11.67 & 10.67 & \textbf{27.67} & 19.67 & \textbf{13.67} & \textbf{8.33} & 67.29 \\
& Llama-3.1-8B-Instruct     & 21.33 & 13.00 & 12.67 & 11.00 & 25.00 & 18.31 & 10.67 & 9.67  & 25.33 & 18.33 & 12.67 & 8.00 & 61.99 \\
& Qwen3-8B-Instruct         & 22.33 & 13.67 & 12.67 & 12.33 & 26.33 & 19.00 & 11.33 & 10.33 & 26.33 & 19.00 & 13.00 & \textbf{8.33} & 64.88 \\
\midrule
\midrule
\multirow{11}{*}{\shortstack[l]{\textbf{Aligned Diffusion}\\\textbf{LLMs (Ours)}}}
& \textbf{LLaDA-1.5} & & & & & & & & & & & & & \\
& \quad $\hookrightarrow$ + only \method-SFT & 29.33 & 16.00 & 15.33 & 14.67 & 28.67 & 20.10 & 13.67 & 11.00 & 23.33 & 20.00 & 13.33 & 7.00 & 70.81 \\
& \quad $\hookrightarrow$ + \method         & 31.00 & \textbf{16.33} & 15.67 & 15.00 & \textbf{30.33} & \textbf{21.14} & \textbf{14.33} & \textbf{11.33} & 24.33 & 21.00 & \textbf{13.67} & 8.00 & 74.04 \\
\cmidrule(lr){2-15}
& \textbf{LLaDA-8B} & & & & & & & & & & & & & \\
& \quad $\hookrightarrow$ + only \method-SFT & 29.67 & 15.67 & 15.33 & 14.33 & 27.67 & 19.89 & 13.33 & 10.67 & 23.67 & 19.67 & 13.33 & 7.33 & 70.19 \\
& \quad $\hookrightarrow$ + \method         & \textbf{31.67} & \textbf{16.33} & \textbf{16.00} & \textbf{16.00} & \textbf{30.33} & 21.07 & 14.00 & 11.00 & 25.67 & \textbf{21.33} & \textbf{13.67} & 8.00 & \textbf{75.02} \\
\cmidrule(lr){2-15}
& \textbf{Dream-7B} & & & & & & & & & & & & & \\
& \quad $\hookrightarrow$ + only \method-SFT & 28.00 & 15.67 & 15.00 & 14.33 & 26.67 & 19.79 & 13.00 & 10.67 & 22.33 & 19.33 & 12.67 & 7.33 & 68.26 \\
& \quad $\hookrightarrow$ + \method         & 30.33 & 16.00 & 15.67 & 15.00 & 29.00 & 20.56 & 13.67 & \textbf{11.33} & 24.00 & 20.33 & 13.33 & 7.67 & 72.30 \\
\cmidrule(lr){2-15}
& \textbf{MMaDA} & & & & & & & & & & & & & \\
& \quad $\hookrightarrow$ + only \method-SFT & 27.00 & 15.33 & 14.67 & 14.00 & 25.67 & 19.24 & 12.67 & 10.33 & 21.33 & 18.67 & 12.67 & 7.33 & 66.30 \\
& \quad $\hookrightarrow$ + \method         & 29.67 & 15.67 & 15.33 & 14.67 & 28.33 & 20.32 & 13.33 & 11.00 & 23.00 & 19.67 & 13.00 & 7.67 & 70.55 \\
\bottomrule
\end{tabular}}
\caption{Fine-grained per-dimension breakdown of the Human Evaluation subset.}
\label{tab:per_dim_evaluation_human}
\end{table*}

\section{Evaluation Rubrics}
\label{app:evaluation-rubrics}

The IIF evaluation rubric is organized around four structural dimensions:
local constraint execution (D1), contextual integration (D2), boundary quality
(D3), and global requirement (D4). These dimensions are applied consistently
across all three benchmark levels. Rather than collapsing output quality into a
single holistic score, this design separates the model's ability to fulfill
in-place constraints from its ability to maintain coherence at the local
transition and global response levels. Each dimension is instantiated with
level-specific sub-criteria that reflect the distinct focal points of L1
(literal fidelity), L2 (semantic realization), and L3 (discourse function
execution). The dimension labels serve as a shared structural scaffold rather
than shared definitions.

Table~\ref{tab:iif-rubric-full} summarizes the per-sub-item point allocation across all levels. Total scores are computed by direct summation ($S = D1 + D2 + D3 + D4$, max 100); the decreasing weight of D4 across levels reflects the increasing complexity of local execution at L2 and L3. For L3-D1, the general point structure is further instantiated by discourse function in Table~\ref{tab:l3-d1-functions}. The score descriptions in Table~\ref{tab:l3-d1-functions} provide approximate reference points to help annotators map real cases to the corresponding level of criterion fulfillment.

\begin{table*}[t]
\centering
\small
\setlength{\tabcolsep}{5pt}
\begin{tabularx}{\textwidth}{
  l l
  >{\raggedright\arraybackslash}X P{0.55cm}
  >{\raggedright\arraybackslash}X P{0.55cm}
  >{\raggedright\arraybackslash}X P{0.55cm}
}
\toprule
\multicolumn{2}{l}{\textbf{Dim / Sub}}
  & \textbf{L1 Criterion}        & \textbf{Pts}
  & \textbf{L2 Criterion}        & \textbf{Pts}
  & \textbf{L3 Criterion}        & \textbf{Pts} \\
\midrule

\multirow{4}{*}{\textbf{D1}}
  & a
    & Verbatim anchor retention
    & 25
    & Style onset sharpness
    & 15
    & Function-specific sub-item a$^\dagger$
    & 15 \\
  & b
    & Positional appropriateness
    & 10
    & Style depth: beyond surface markers
    & 15
    & Function-specific sub-item b$^\dagger$
    & 15 \\
  & c
    & No paraphrastic substitution
    & 5
    & Residual style interference
    & 10
    & Function-specific sub-item c$^\dagger$
    & 10 \\
\cmidrule(lr){2-8}
  & \textit{Total}
    & & \textit{40}
    & & \textit{40}
    & & \textit{40} \\
\midrule

\multirow{4}{*}{\textbf{D2}}
  & a
    & Syntactic continuity (Pre)
    & 8
    & Entity consistency (within span)
    & 10
    & Argumentative consistency
    & 12 \\
  & b
    & Semantic flow (Pre)
    & 8
    & Thematic continuity
    & 10
    & Logical placement
    & 10 \\
  & c
    & No pre-constraint redundancy
    & 4
    & No internal hallucination
    & 5
    & Logical extensibility
    & 8 \\
\cmidrule(lr){2-8}
  & \textit{Total}
    & & \textit{20}
    & & \textit{25}
    & & \textit{30} \\
\midrule

\multirow{4}{*}{\textbf{D3}}
  & a
    & Syntactic continuation (Post)
    & 8
    & Stylistic boundary shift
    & 10
    & Clear function onset signal
    & 10 \\
  & b
    & Semantic continuation (Post)
    & 8
    & No meta-commentary (cap rule)
    & 10
    & No function anticipation
    & 10 \\
  & c
    & No immediate repetition
    & 4
    & —
    & —
    & —
    & — \\
\cmidrule(lr){2-8}
  & \textit{Total}
    & & \textit{20}
    & & \textit{20}
    & & \textit{20} \\
\midrule

\multirow{4}{*}{\textbf{D4}}
  & a
    & Instruction relevance
    & 8
    & Inter-section narrative unity
    & 8
    & Grammatical correctness
    & 5 \\
  & b
    & No truncation (output completeness)
    & 6
    & Global task fulfillment
    & 7
    & Clarity of expression
    & 5 \\
  & c
    & No verbatim input copying
    & 6
    & —
    & —
    & —
    & — \\
\cmidrule(lr){2-8}
  & \textit{Total}
    & & \textit{20}
    & & \textit{15}
    & & \textit{10} \\
\midrule

\multicolumn{2}{l}{\textbf{Grand Total}}
  & & \textbf{100}
  & & \textbf{100}
  & & \textbf{100} \\
\bottomrule
\end{tabularx}
\caption{IIF evaluation rubric: per-sub-item point allocation. D1--D4 serve as structural labels for level-specific criteria.}
\label{tab:iif-rubric-full}

\smallskip
\noindent\footnotesize
$^\dagger$ L3-D1 criteria vary by function (e.g., challenge strength for \textsc{Counterexample}; testability for \textsc{Hypothesis}), see Table~\ref{tab:l3-d1-functions}. \\
\end{table*}

\begin{table*}[t]
\centering
\small
\setlength{\tabcolsep}{4pt}
\begin{tabularx}{\textwidth}{
  L{2.6cm}
  >{\raggedright\arraybackslash}X
  >{\raggedright\arraybackslash}X
  >{\raggedright\arraybackslash}X
}
\toprule
\textbf{Function}
& \textbf{D1a (15 pts)}
& \textbf{D1b (15 pts)}
& \textbf{D1c (10 pts)} \\
\midrule

\textsc{Counterexample}
& A specific, tangible scenario or case is presented, rather than an abstract statement.
& The example genuinely complicates or contradicts the main argument in the pre-constraint text.
& The example satisfies any source-domain condition specified in the constraint. For instance, if ``from everyday life'' is specified, the example is drawn from common experience rather than a technical or academic domain. If no such condition is specified, the example should remain contextually appropriate and concrete. \\

\midrule

\textsc{Hypothesis}
& Conditional or speculative framing is present, such as ``if'', ``suppose'', ``imagine'', or ``were it the case''.
& The hypothesis makes a claim that could in principle be examined or argued for/against.
& The hypothesis is introduced as an open question to be examined, not as a settled conclusion. \\

\midrule

\textsc{Summary}
& The summary covers the key points made in the pre-constraint text without omitting critical arguments.
& The summary is noticeably more concise than the content it summarizes.
& The summary does not introduce new claims. The related L3-D2c criterion, logical extensibility, is evaluated using the same evidence, but its 8 points remain counted under D2 to preserve the 100-point total. \\

\midrule

\textsc{Transition}
& The content of the two ideas being bridged is genuinely connected, not merely juxtaposed. The transition explains why or how the second idea follows from the first, not just that it does. Score 15 if the logical link is explicit and substantive; 8 if implicit but traceable; 0 if the two ideas remain essentially unrelated despite surface connectives.
& Surface connective language is present, such as ``however'', ``building on this'', ``turning now to'', or ``this leads us to''. Score 15 if explicit connectives are present; 8 if implicit but detectable; 0 if absent.
& The transition is appropriately brief, not a paragraph-long digression. Score 10 if concise and functional; 5 if slightly extended; 0 if excessively long. \\

\midrule

\textsc{Analogy}
& The analogy draws from a domain clearly different from the main topic.
& The structural mapping between the analogy domain and the main argument is logically valid, such that the shared structure genuinely holds.
& The analogy is self-explanatory to a general reader without additional context. Score 10 if immediately clear; 5 if it requires minor inferential effort; 0 if opaque. \\

\midrule

\textsc{Rebuttal}
& An opposing viewpoint or objection is clearly identified.
& The refutation provides a reason or evidence against the opposing view, not merely an assertion.
& The opposing view is presented fairly and is not a strawman. \\

\midrule

\textsc{Parallelism}
& Two or more clauses or sentences share the same syntactic pattern. Score 15 for ${\geq}3$ parallel structures; 10 for exactly 2; 0 for none.
& Each parallel element introduces distinct semantic content rather than synonymous repetition.
& The parallel structure creates a sense of emphasis, rhythm, or cumulative rhetorical force. \\

\midrule

\textsc{Rhetorical Question}
& At least one grammatically correct question is present in the constrained section. Score 15 if present; 0 if absent.
& Conditional on D1a $> 0$: the question is not immediately answered and is posed for rhetorical effect. If D1a $= 0$, D1b is automatically scored 0. Otherwise, score 15 if fully rhetorical; 8 if partially answered; 0 if directly and completely answered.
& The question advances or emphasizes the main argument; it is not tangential or unrelated to the discourse. \\

\bottomrule
\end{tabularx}
\caption{Function-specific instantiations of L3-D1. All L3 discourse functions share the same 15/15/10 point structure, but the concrete execution criteria are defined separately for each function.}
\label{tab:l3-d1-functions}
\end{table*}

For L2 specifically, the forced insertion of a local style may introduce an
inherent tension with the surrounding global context. A model output can
therefore appear less contextually smooth not because it fails to follow the
instruction, but because the assigned local style itself conflicts with the
preceding context. To account for this local--global tension, we apply a
conflict-aware adjustment to the L2-D2 score.

For each L2 instance, annotators first assign a raw contextual-integration
score $S_{\text{raw},i}$ for D2. They also independently estimate a
ConflictScore, which measures the intrinsic difficulty of harmonizing the
assigned local style with the surrounding context. The ConflictScore aggregates
three sources of tension: register distance, emotional-valence conflict, and
structural incompatibility. Table~\ref{tab:iif-conflict} summarizes the three
components. Formally,
\[
\text{ConflictScore}_i = C_{1,i} + C_{2,i} + C_{3,i},
\]
with a maximum value of 100.

We then compute a conflict-adjusted D2 score for each L2 instance:
\[
S_{\text{adj},i}
=
S_{\text{raw},i}
+
(25 - S_{\text{raw},i})
\cdot
\frac{\text{ConflictScore}_i}{100}
\cdot
\beta,
\]
where $S_{\text{D2-raw},i}$ is the raw D2 score of instance $i$ and 25 is
the maximum possible score for the L2-D2 dimension and
$\beta \in [0,1]$ controls the strength of the adjustment.  The
adjustment is compensatory rather than punitive: a larger ConflictScore
indicates a stronger intrinsic mismatch between the assigned local style and
the surrounding context, and therefore reduces the penalty imposed by a low
raw D2 score. In other words, high-conflict L2 instances are not treated as
equally easy contextual-integration cases as low-conflict ones.

The L2-D2 value reported in our main results is the mean adjusted score
$\overline{S}_{\text{adj}}$ over all L2 instances. Consequently, unlike the
raw rubric scores, the reported L2-D2 value is not restricted to the discrete
annotation grid. The overall L2 score is computed as
\[
S_{\text{L2}}
=
D1_{\text{raw}}
+
\overline{S}_{\text{adj}}
+
D3_{\text{raw}}
+
D4_{\text{raw}}.
\]

\section{Case Study}
\label{app:case-study}

\paragraph{Case: Style Constraint}

This case illustrates a style-control failure. The requested continuation should maintain a joyful pastoral style, bright with sunlight, birdsong, and dancing streams. The \method answer follows this style more consistently from beginning to end, using images such as green fields, warm sunlight, birdsong, and streams. The Vanilla answer partially recovers in the second half, but its opening relies on twilight, shadows, and moonlight, which conflicts with the requested brightness and pastoral joy.

\begin{figure*}[tp]
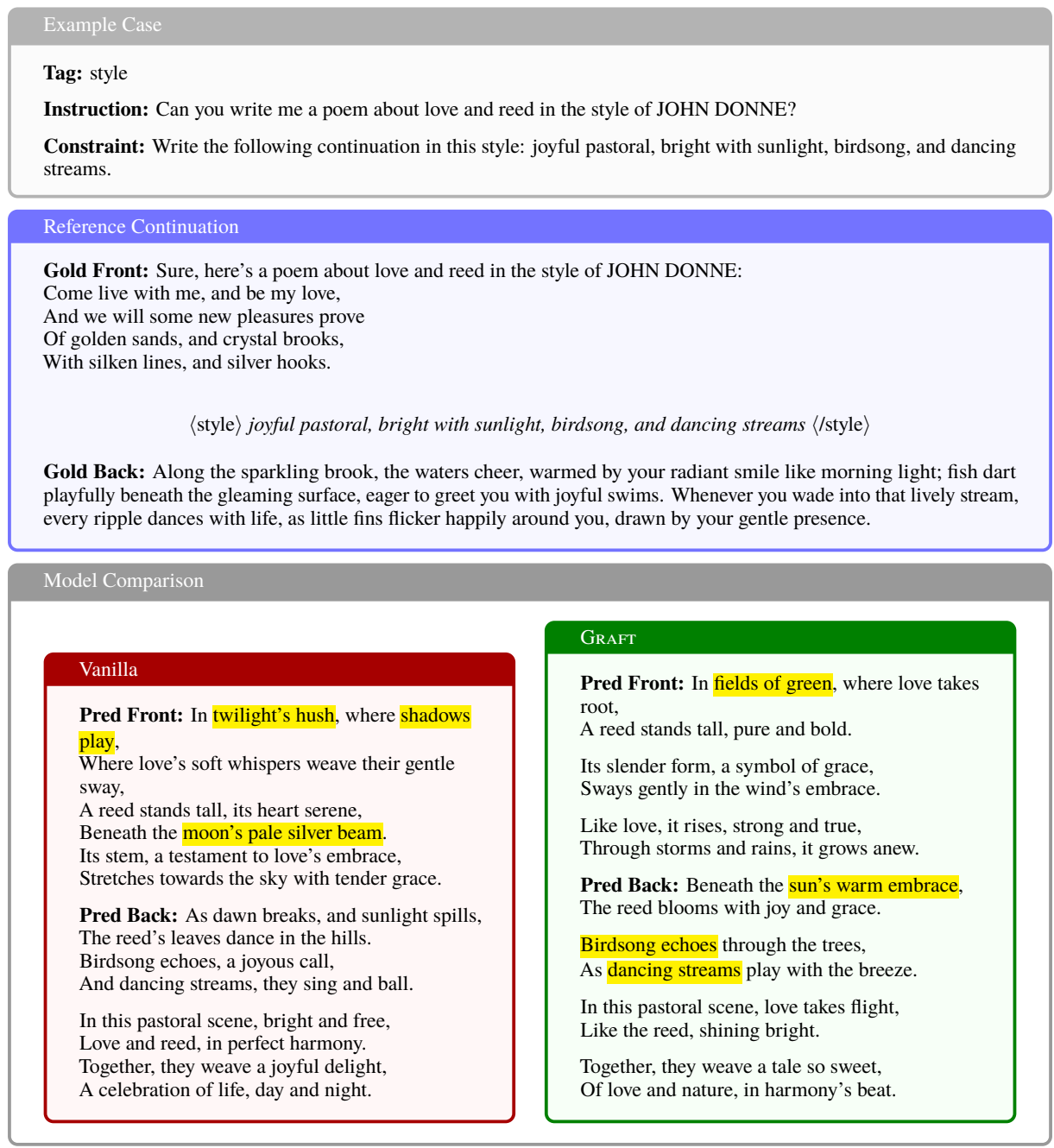

\centering
\small

\begin{tcolorbox}[width=\textwidth, title=Example Case, colback=gray!3, colframe=gray!60]
\textbf{Tag:} style

\medskip

\textbf{Instruction:} Can you write me a poem about love and reed in the style of JOHN DONNE?

\medskip

\textbf{Constraint:} Write the following continuation in this style: joyful pastoral, bright with sunlight, birdsong, and dancing streams.
\end{tcolorbox}

\begin{tcolorbox}[width=\textwidth, title=Reference Continuation, colback=blue!3, colframe=blue!55]
\textbf{Gold Front:}
Sure, here's a poem about love and reed in the style of JOHN DONNE:

Come live with me, and be my love,\\
And we will some new pleasures prove\\
Of golden sands, and crystal brooks,\\
With silken lines, and silver hooks.

\medskip

\[
\langle\text{style}\rangle\ 
\textit{joyful pastoral, bright with sunlight, birdsong, and dancing streams}
\ \langle\text{/style}\rangle
\]

\textbf{Gold Back:}
Along the sparkling brook, the waters cheer, warmed by your radiant smile like morning light; fish dart playfully beneath the gleaming surface, eager to greet you with joyful swims. Whenever you wade into that lively stream, every ripple dances with life, as little fins flicker happily around you, drawn by your gentle presence.
\end{tcolorbox}

\begin{tcolorbox}[width=\textwidth, title=Model Comparison, colback=white, colframe=black!40]

\noindent
\begin{minipage}[t]{0.485\textwidth}
\begin{tcolorbox}[
  title=Vanilla,
  colback=red!3,
  colframe=red!65!black,
  before upper={\raggedright}
]
\textbf{Pred Front:}
In \hl{twilight's hush}, where \hl{shadows play},\\
Where love's soft whispers weave their gentle sway,\\
A reed stands tall, its heart serene,\\
Beneath the \hl{moon's pale silver beam}.\\
Its stem, a testament to love's embrace,\\
Stretches towards the sky with tender grace.

\medskip

\textbf{Pred Back:}
As dawn breaks, and sunlight spills,\\
The reed's leaves dance in the hills.\\
Birdsong echoes, a joyous call,\\
And dancing streams, they sing and ball.

\medskip

In this pastoral scene, bright and free,\\
Love and reed, in perfect harmony.\\
Together, they weave a joyful delight,\\
A celebration of life, day and night.
\end{tcolorbox}
\end{minipage}
\hfill
\begin{minipage}[t]{0.485\textwidth}
\begin{tcolorbox}[
  title=\method,
  colback=green!3,
  colframe=green!50!black,
  before upper={\raggedright}
]
\textbf{Pred Front:}
In \hl{fields of green}, where love takes root,\\
A reed stands tall, pure and bold.

\medskip

Its slender form, a symbol of grace,\\
Sways gently in the wind's embrace.

\medskip

Like love, it rises, strong and true,\\
Through storms and rains, it grows anew.

\medskip

\textbf{Pred Back:}
Beneath the \hl{sun's warm embrace},\\
The reed blooms with joy and grace.

\medskip

\hl{Birdsong echoes} through the trees,\\
As \hl{dancing streams} play with the breeze.

\medskip

In this pastoral scene, love takes flight,\\
Like the reed, shining bright.

\medskip

Together, they weave a tale so sweet,\\
Of love and nature, in harmony's beat.
\end{tcolorbox}
\end{minipage}

\end{tcolorbox}

\caption{Style constraint case. The Vanilla answer conflicts with the requested bright pastoral tone, while the \method answer maintains the target style more consistently.}
\end{figure*}

\paragraph{Case: Discourse Constraint}

This case illustrates a discourse-control failure. The requested continuation should name and contest an implicit assumption. The \method answer identifies a plausible assumption: that image formation depends only on the interaction between the electron beam and the sample. It then contests this assumption by introducing electron optics and detector systems as necessary parts of the imaging pipeline. The Vanilla answer also attempts to contest an assumption, but its reference to gravity and pressure is distracting and technically weak in this context, making it less coherent as a continuation.

\begin{figure*}[tp]
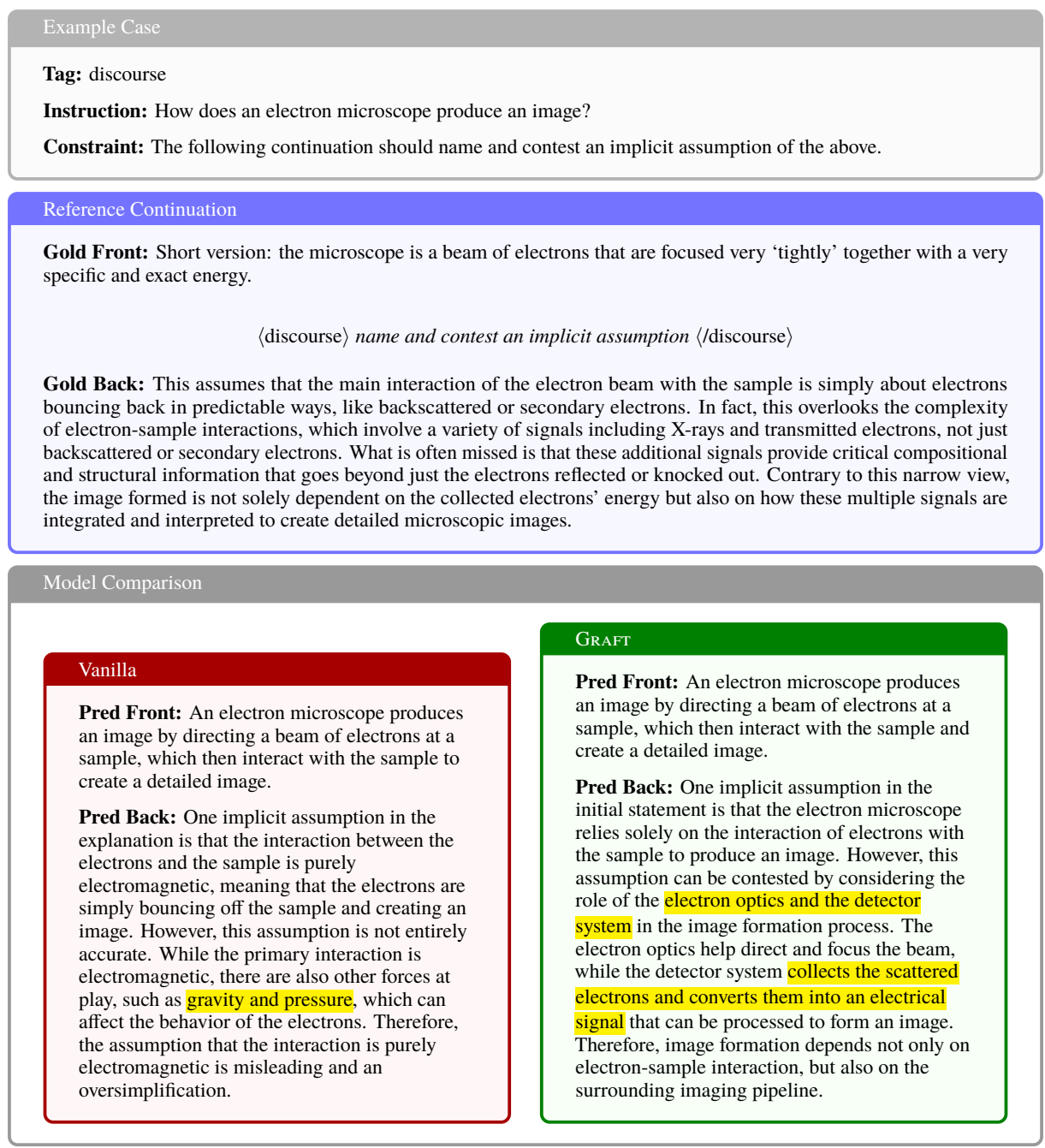

\centering
\small

\begin{tcolorbox}[width=\textwidth, title=Example Case, colback=gray!3, colframe=gray!60]
\textbf{Tag:} discourse

\medskip

\textbf{Instruction:} How does an electron microscope produce an image?

\medskip

\textbf{Constraint:} The following continuation should name and contest an implicit assumption of the above.
\end{tcolorbox}

\begin{tcolorbox}[width=\textwidth, title=Reference Continuation, colback=blue!3, colframe=blue!55]
\textbf{Gold Front:}
Short version: the microscope is a beam of electrons that are focused very `tightly' together with a very specific and exact energy.

\medskip

\[
\langle\text{discourse}\rangle\ 
\textit{name and contest an implicit assumption}
\ \langle\text{/discourse}\rangle
\]

\textbf{Gold Back:}
This assumes that the main interaction of the electron beam with the sample is simply about electrons bouncing back in predictable ways, like backscattered or secondary electrons. In fact, this overlooks the complexity of electron-sample interactions, which involve a variety of signals including X-rays and transmitted electrons, not just backscattered or secondary electrons. What is often missed is that these additional signals provide critical compositional and structural information that goes beyond just the electrons reflected or knocked out. Contrary to this narrow view, the image formed is not solely dependent on the collected electrons' energy but also on how these multiple signals are integrated and interpreted to create detailed microscopic images.
\end{tcolorbox}

\begin{tcolorbox}[width=\textwidth, title=Model Comparison, colback=white, colframe=black!40]

\noindent
\begin{minipage}[t]{0.485\textwidth}
\begin{tcolorbox}[
  title=Vanilla,
  colback=red!3,
  colframe=red!65!black,
  before upper={\raggedright}
]
\textbf{Pred Front:}
An electron microscope produces an image by directing a beam of electrons at a sample, which then interact with the sample to create a detailed image.

\medskip

\textbf{Pred Back:}
One implicit assumption in the explanation is that the interaction between the electrons and the sample is purely electromagnetic, meaning that the electrons are simply bouncing off the sample and creating an image. However, this assumption is not entirely accurate. While the primary interaction is electromagnetic, there are also other forces at play, such as \hl{gravity and pressure}, which can affect the behavior of the electrons. Therefore, the assumption that the interaction is purely electromagnetic is misleading and an oversimplification.
\end{tcolorbox}
\end{minipage}
\hfill
\begin{minipage}[t]{0.485\textwidth}
\begin{tcolorbox}[
  title=\method,
  colback=green!3,
  colframe=green!50!black,
  before upper={\raggedright}
]
\textbf{Pred Front:}
An electron microscope produces an image by directing a beam of electrons at a sample, which then interact with the sample and create a detailed image.

\medskip

\textbf{Pred Back:}
One implicit assumption in the initial statement is that the electron microscope relies solely on the interaction of electrons with the sample to produce an image. However, this assumption can be contested by considering the role of the \hl{electron optics and the detector system} in the image formation process. The electron optics help direct and focus the beam, while the detector system \hl{collects the scattered electrons and converts them into an electrical signal} that can be processed to form an image. Therefore, image formation depends not only on electron-sample interaction, but also on the surrounding imaging pipeline.
\end{tcolorbox}
\end{minipage}

\end{tcolorbox}

\caption{Discourse constraint case. The Vanilla answer contests a weak and distracting assumption, while the \method answer identifies a more relevant implicit assumption and contests it coherently.}
\end{figure*}

\end{document}